\documentclass[sigconf]{acmart}
\usepackage{enumitem}
\usepackage{graphicx}
\usepackage{multicol}
\usepackage{subcaption}
\usepackage{xcolor}
\usepackage{bm}
\usepackage{wrapfig}
\usepackage{makecell}
\usepackage[subtle]{savetrees}
\usepackage{cleveref}
\usepackage{tikz}

\definecolor{diff}{RGB}{0,0,0}
\definecolor{diff2}{RGB}{0,0,0}
\definecolor{diff3}{RGB}{0,0,0}

\copyrightyear{2022}
\acmYear{2022}
\setcopyright{acmlicensed}
\acmConference[SIGMOD '22] {Proceedings of the 2022 International Conference on Management of Data}{June 12--17, 2022}{Philadelphia, PA, USA.}
\acmBooktitle{Proceedings of the 2022 International Conference on Management of Data (SIGMOD '22), June 12--17, 2022, Philadelphia, PA, USA}
\acmPrice{15.00}
\acmISBN{978-1-4503-9249-5/22/06}
\acmDOI{10.1145/3514221.3517848}
\begin{document}
\fancyhead{}
\title[Hidden Trade-Offs in Deep Learning Preprocessing Pipelines]{Where Is My Training Bottleneck?\\ Hidden Trade-Offs in Deep Learning Preprocessing Pipelines}

\author{Alexander Isenko,\; Ruben Mayer,\; Jeffrey Jedele}
\affiliation{%
  \institution{Technical University of Munich, Germany}
  \streetaddress{}
  \city{}
  \state{}
  \country{}
  \postcode{}
}
\email{{alex.isenko}, {ruben.mayer}, {jeffrey.jedele}@tum.de}

\author{Hans-Arno Jacobsen}
\affiliation{%
  \institution{University of Toronto, Canada}
  \streetaddress{}
  \city{}
  \state{}
  \country{}
  \postcode{}
}
\email{jacobsen@eecg.toronto.edu}

\renewcommand{\shortauthors}{Isenko, et al.}

\begin{abstract}
Preprocessing pipelines in deep learning aim to provide sufficient data throughput to keep the training processes busy.
Maximizing resource utilization is becoming more challenging as the throughput of training processes increases with hardware innovations (e.g., faster GPUs, TPUs, and inter-connects) and advanced parallelization techniques that yield better scalability.
At the same time, the amount of training data needed in order to train increasingly complex models is growing.
As a consequence of this development, data preprocessing and provisioning are becoming a severe bottleneck in end-to-end deep learning pipelines.

In this paper, we provide an in-depth analysis of data preprocessing pipelines from four different machine learning domains.
We introduce a new perspective on efficiently preparing datasets for end-to-end deep learning pipelines and extract individual trade-offs to optimize throughput, preprocessing time, and storage consumption.
Additionally, we provide an open-source profiling library that can automatically decide on a suitable preprocessing strategy {\color{diff}to maximize throughput}.
By applying our generated insights to real-world use-cases, we obtain an increased throughput of {\color{diff}3$\times$ to 13$\times$}  compared to an untuned system while keeping the pipeline functionally identical.
These findings show the enormous potential of data pipeline tuning.
\end{abstract}




\maketitle
\section{Introduction}

\begin{tikzpicture}
\begin{scope}[overlay]
\node[text width=17.5cm] at ([yshift=-20cm,xshift=-11cm]current page.south) {(c) Owner 2022. This is the authors' version of the work. It is posted here for your personal use. Not for redistribution. \newline The definitive version is published in Proceedings of the 2022 International Conference on Management of Data (SIGMOD '22), June 12--17, 2022, Philadelphia, PA, USA. https://doi.org/10.1145/3514221.3517848.};
\end{scope}
\end{tikzpicture}

\label{sec:introduction}
\vspace{-0.4cm}
Deep learning (DL) models are used in multiple areas, ranging from e-mail spam filtering~\cite{blanzieri2008survey} in natural language processing (NLP) to image segmentation tasks for autonomous driving~\cite{he2017mask} in computer vision (CV).
The improvement of these models is not only based on more advanced architectures or algorithms~\cite{vinyals2015show, szegedy2015going, lai2015simultaneous}, but also on increased quality and quantity of training data~\cite{Krizhevsky:2012:ICD:2999134.2999257,deng2009imagenet,lin2014microsoft,everingham2011pascal,szegedy2015going}.
Having more training data usually proves to be beneficial to the model performance~\cite{banko2001scaling}.

\vspace{-0.3cm}
\begin{figure}[h]
    \centering
    \includegraphics[width=0.47\textwidth]{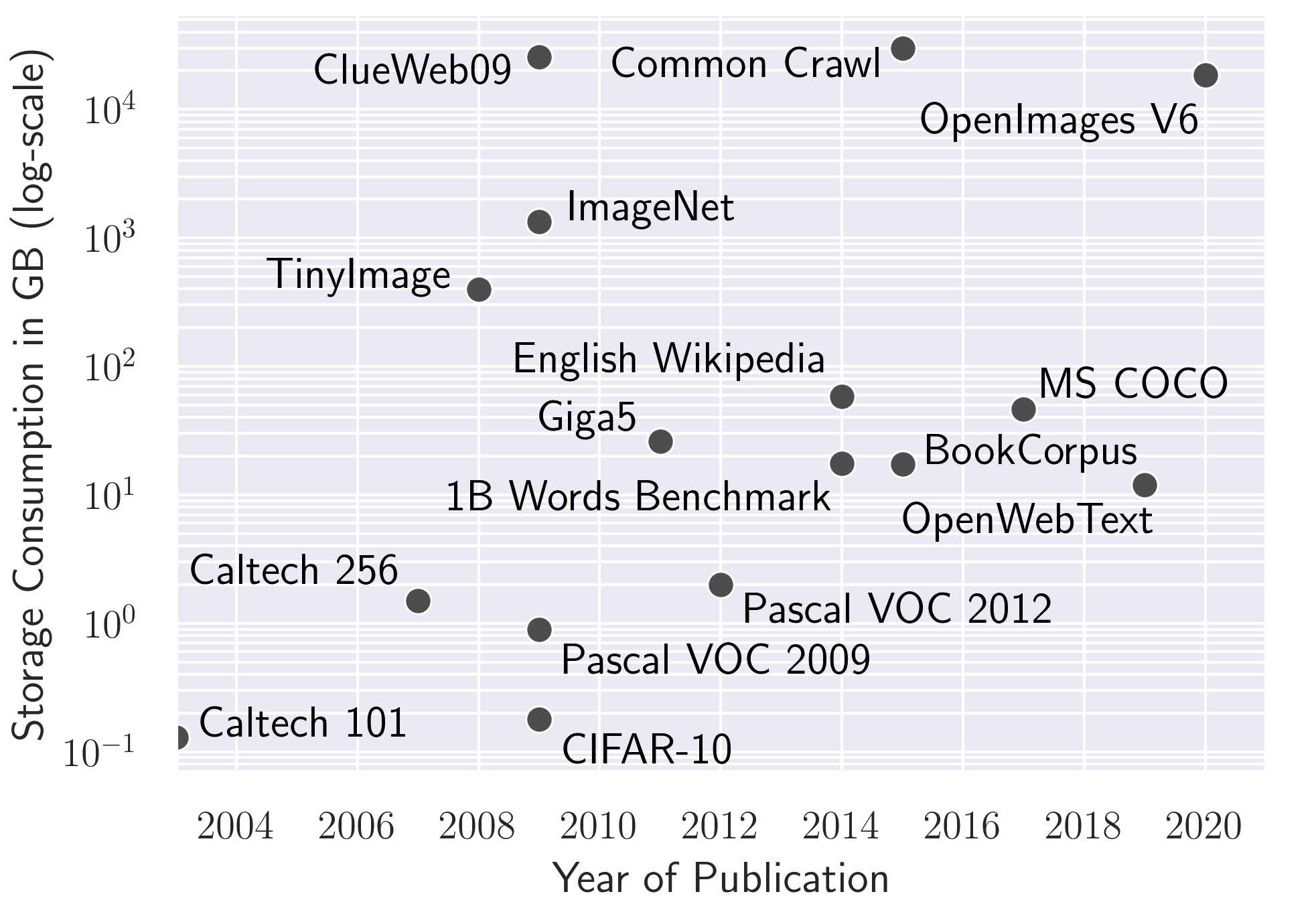}
    \vspace{-0.3cm}
    \caption{Storage consumption of real-world CV and NLP datasets over time on a logarithmic scale. CV:~\cite{fei2004learning, griffin2007caltech, torralba200880, deng2009imagenet, pascal-voc-2009, everingham2011pascal, krizhevsky2009learning, lin2014microsoft, OpenImages2}, NLP:~\cite{callan2009clueweb09, giga5,chelba2013billion, engwiki, zhu2015aligning, commoncrawl, Gokaslan2019OpenWeb}.}
    \label{fig:all-datasets}
\end{figure}
\vspace{-0.2cm}

The process to train a DL model consists of repeatedly iterating over the entire training dataset, measuring up to hundreds of iterations depending on the task at hand and the model complexity~\cite{peters2018deep,he2016deep,szegedy2015rethinking,radford2018improving}.
Popular datasets show exponential storage consumption increase over time (Fig.~\ref{fig:all-datasets}), which makes data preprocessing harder, as local processing is not viable anymore due to memory limitations. Both distributed storage solutions~\cite{weil2006ceph} as well as distributed processing~\cite{gabriel2004open, vishnu2016distributed, beam,zaharia2010spark, 10.1145/3363554} can lead to new difficulties with network I/O and latency, which makes data preprocessing an integral part of the end-to-end DL pipeline.
A Google study on their cluster fleet showed that the preprocessing pipeline takes more than a third of the total preprocessing time for 20\% of their jobs~\cite{murray2021tf}.

\begin{figure}[h]
    \centering
    \includegraphics[width=0.48\textwidth]{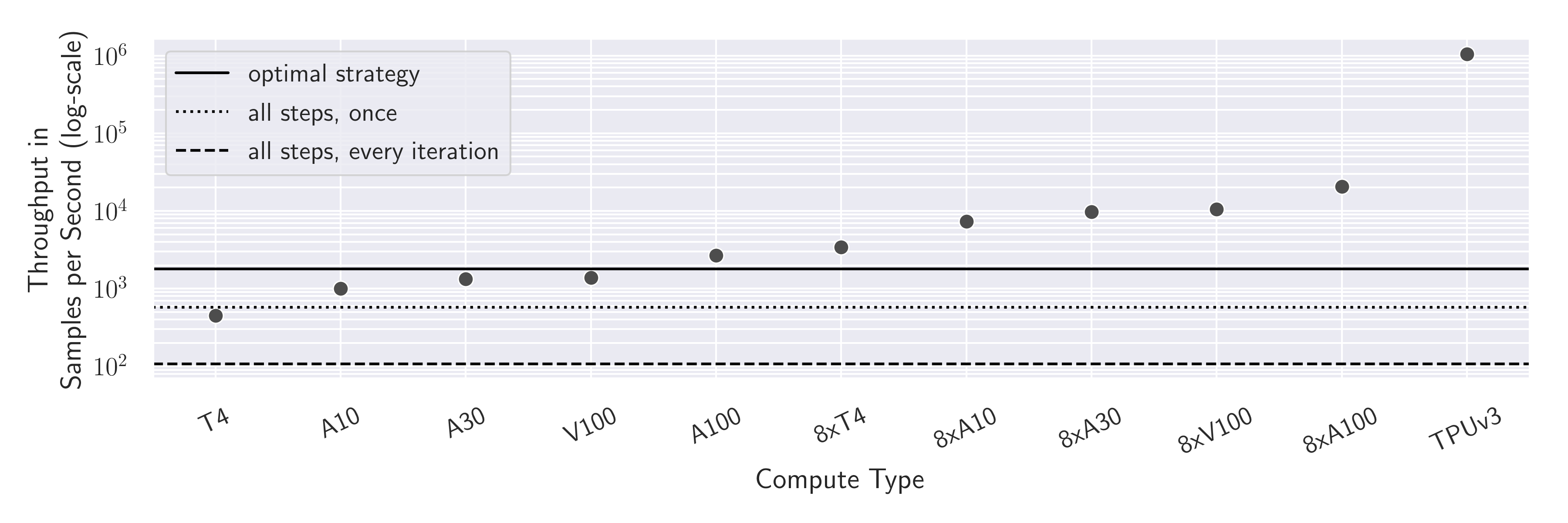}
    \vspace{-0.8cm}
    \caption{Throughput of ResNet-50~\cite{he2016deep} for different hardware configurations. Black lines show the preprocessing throughput for the different strategies from Tab.~\ref{tab:intro:trade-offs-cv}. GPU profiling data from NVIDIA~\cite{nvidiabenchmarks2020} and TPUv3 by Ying et al.~\cite{ying2018image}.}
    \label{fig:gpu-resnet-throughput}
\end{figure}

Optimizing the model training is an active research topic that focuses on decreasing the total training time and increasing the data ingestion rate of the training process.
There are many methods on training performance optimization and model optimization for both single GPU~\cite{micikevicius2017mixed,vanholder2016efficient,han2015deep} and multi-GPU setups~\cite{jager2018parallelized,sergeev2018horovod,Jia2019,ren2019performance} which allow for horizontal scaling with more hardware~\cite{nvidiabenchmarks2020}.
Recent hardware innovations help with improved model performance (cf. Fig.~\ref{fig:gpu-resnet-throughput}).
Therefore, it is essential to optimize preprocessing pipelines to keep up with the training process speed.

The preprocessing part of a DL pipeline consists of multiple successive data transformation steps applied to the initial dataset until the final data representation matches the model input dimensions.
This transformation can be performed once before the training or in every iteration while the training is happening.
For example, CV pipelines from DL models that established new landmarks at their respective times~\cite{krizhevsky2012imagenet,zeiler2013visualizing,Simonyan15,he2016deep,szegedy2015rethinking} follow a common pattern of preprocessing steps: \textit{read} the image from a storage device, \textit{decode} it into an RGB matrix and \textit{resize} the image to fit the model input dimensions. 
These steps can be followed by data augmentation, e.g., \textit{pixel-centering}, \textit{random-cropping} or a \textit{rotation}, depending on the particular use-case.
Preprocessing the full dataset once before training is viable if one wants to avoid the processing overhead in every iteration.

However, the final data representation and the storage device can negatively affect the preprocessing throughput.
The training process's data ingestion can be throttled by I/O bottlenecks when loading the preprocessed data.
The storage consumption typically increases at later preprocessing steps, as the corresponding data representations often store data inefficiently to facilitate processing (e.g., JPG~\cite{wallace1992jpeg} vs. an RBG matrix).
This additional storage consumption can be a determining factor that slows down the final throughput. The file system, storage device, and the data loader from the DL framework may not be able to read the data fast enough (cf. Section~\ref{sec:analysis}).

We propose a new, more flexible way to look at DL preprocessing pipelines, where the decision for \textbf{each} preprocessing step to apply it \textit{once} or in \textit{every iteration} can be made freely based on quantifiable trade-offs.
Such quantification can be provided by profiling.

\begin{figure}[h]
    \centering
    \includegraphics[width=0.47\textwidth]{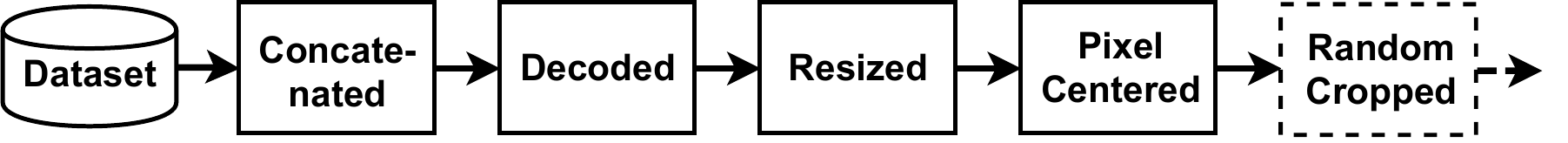}
    \vspace*{-0.2cm}
    \caption{CV preprocessing pipeline}
    \label{fig:cv-pipeline}
\end{figure}
\vspace{-0.3cm}

To motivate this new view on preprocessing pipelines, we performed experiments using different configurations of a CV pipeline (Fig.~\ref{fig:cv-pipeline}){\color{diff} \footnote{The only step which has to be applied every iteration is \textit{random-crop}, as it is not deterministic (dotted line).}}.
{\color{diff} Performing all preprocessing steps \textit{at once} increases the throughput by $5.4\times$ compared to \textit{at every iteration} (Tab.~\ref{tab:intro:trade-offs-cv}).
However, this increases storage consumption by more than $10\times$.
In contrast, preprocessing the dataset \textit{once} just until the resize step results in a $16.7\times$ throughput increase while increasing the storage consumption only by $3.4\times$ compared to processing all steps at every iteration.}

\begin{table}[h]
\scalebox{0.73}{
\begin{tabular}{l|r|r}
\textbf{Preprocessing strategy}        & \textbf{Throughput in} $\frac{samples}{s}$ & \textbf{Storage Consumption in GB}     \\ \hline
all steps at \textit{every iteration}       & {\color{diff}107}                 & \textbf{146}                  \\ \hline
all steps \textit{once}                     & {\color{diff}576}                 & 1535                          \\ \hline
until \textit{resize} step, \textit{once}  & {\color{diff}\textbf{1789}}   & 494                           \\
\end{tabular}
}
\caption{Trade-offs for the CV pipeline at different preprocessing strategies.}
\label{tab:intro:trade-offs-cv}
\end{table}
\vspace{-0.5cm}

When comparing the data processing rate of a popular CV model, ResNet-50, to the different preprocessing strategies on state-of-the-art GPUs, we see that stalls on the A10, A30, and V100 can be prevented by using the optimal strategy (Fig. ~\ref{fig:gpu-resnet-throughput}).
In multi-node training setups and when using specialized hardware (TPUs), increasing preprocessing throughput demands becomes even more evident.

The idea of opening up the preprocessing pipeline and the resulting trade-offs have not been explored yet in a comprehensive fashion.
This void prevents ML practitioners from optimizing their end-to-end DL pipelines.
They lack guidance in how to do that, as well as tooling support to automate such optimizations.

\vspace{-0.5cm}
\vspace{-0.4cm}

In this paper, we close this research gap by performing a comprehensive analysis of preprocessing pipelines from a broad range of different ML domains.
In doing so, we present practical insights into the pipelines themselves as well as the methodologies to analyze bottlenecks and an automated tool to perform profiling of arbitrary pipelines.
This opens up a new dimension in end-to-end ML system optimizations, which was not considered in prior works that targeted the pipeline optimization with respect to the model accuracy~\cite{mohan2020analyzing, kang2020jointly}.

Our contributions are:
\begin{enumerate}
    \item \textbf{We profile seven different real-world pipelines} and define the trade-offs and characteristics that allow practitioners to improve existing pipelines by optimizing at the location with the greatest impact on the training throughput. This way, we could improve training throughput by up to {\color{diff}3-13$\times$ compared to fully preprocessing once.}
    \item We provide lessons learned, where we \textbf{summarize the problems and unexpected findings} we encountered that can limit pipeline throughput. For example, we found that storage consumption {\color{diff}can affect the throughput negatively in different ways.} These insights can be used to clear up common misconceptions, and practitioners can be more aware of the impact the preprocessing pipeline has on the training performance.
    \item We present an open-source \textbf{profiling library} that automates the decision of which preprocessing strategy to pick based on a user-defined cost model.
\end{enumerate}

Our paper is organized as follows.
In Section~\ref{sec:preprocessing}, we introduce a general model and terminology of preprocessing pipelines.
The experimental setup and the library design are explained in Section~\ref{sec:experiments}.
We present our pipeline analysis and our findings in Section~\ref{sec:analysis}.
Our derived insights are summarized in Section~\ref{sec:lessons-learned}.
Related work is reviewed in Section~\ref{sec:related-work}.
Other approaches for pipeline optimizations are discussed in Section~\ref{sec:discussion} and we conclude the paper in Section~\ref{sec:conclusion}.
\section{Preprocessing Pipelines}
\label{sec:preprocessing}
We begin by describing the set of problems one faces when preparing a dataset for the training process.
This includes determining the hardware requirements, the decision of \textit{where} to preprocess the data, and \textit{when} to preprocess, both of which decisions affect the training throughput in multiple ways.

The preprocessing pipeline can be split into steps which are ran \textit{once} ($S_{\textbf{1}}-S_{\textbf{m}}$), called ``offline'' henceforth, and steps which are performed \textit{every iteration} ($S_{\textbf{m+1}}-S_{\textbf{n}}$), called ``online''.
The set of preprocessing steps depends on the dataset and the model input, but generally, any transformation is a step, like cropping an image or encoding a word.
A preprocessing \textit{strategy} is processing up to (and including) a step offline, and the remainder of the pipeline is executed online. Such a split is accomplished by inserting a \textit{save} step which encodes and writes the data to disk (after $S_\textbf{m}$), and a \textit{load} step that reads that data from disk (before $S_{\textbf{m+1}}$).

\begin{figure}[h]
  \includegraphics[width=0.4\textwidth]{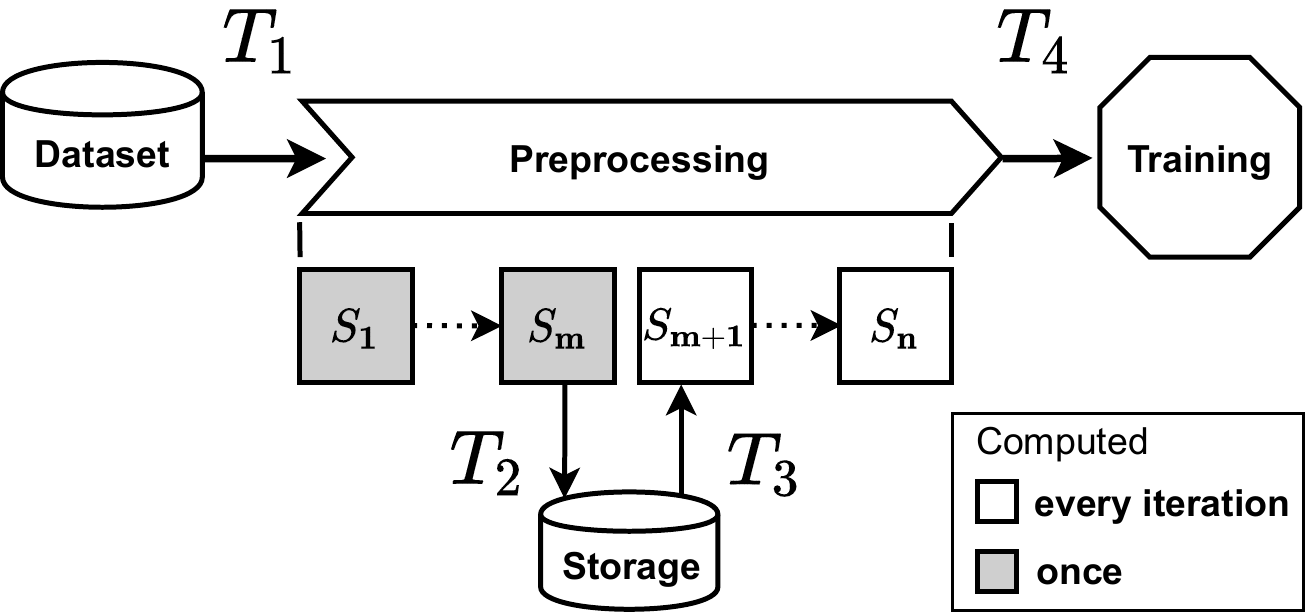}
  \vspace{-0.1cm}
  \caption{General DL preprocessing setup with different theoretical throughputs $T_{1}-T_{4}$ between the preprocessing steps $S_{\textbf{1}}-S_\textbf{n}$.}
  \label{fig:generic-dl-pipeline}
\end{figure}
\vspace{-0.2cm}

We conceptually divide the entire DL training process into three parts - the unprocessed dataset, the preprocessing pipeline, and the training (Fig.~\ref{fig:generic-dl-pipeline}).
They are all connected by four theoretical throughputs $(T_{1}-T_{4})$, which can become processing bottlenecks.

\begin{description}[align=left]
\item[$T_1$] is the read throughput from the dataset to the processing units which handle the \textit{offline} preprocessing.
This throughput is determined by hardware capabilities, such as storage devices, interconnects, processing capabilities, and software capabilities, such as file systems or DL frameworks.

\item[$T_2$] is the write throughput from the offline computed preprocessing step(s) ($S_{\textbf{1}}-S_{\textbf{m}}$) to a storage node.
It is dependent on the throughput of each step, the interconnect to the storage node, and its write speed.

\item[$T_3$] is the read throughput from the storage node to the processing units which handle the \textit{online} preprocessing ($S_{\textbf{m+1}}-S_{\textbf{n}}$) and is subject to the same restrictions as $T_1$.

\item[$T_4$] is the final preprocessing throughput when the data is ready to be fed into the training process.
It is restricted by $T_{3}$, the online step(s) performance, and the interconnect to the training process.
As $T_4$ limits the achievable training throughput, it is the most important to optimize.
\end{description}

In practice, it is often impossible to know the actual performance of future DL models or preprocessing pipelines. 
Only partial benchmarks are available to approximate the training throughput of popular DL models~\cite{nvidiabenchmarks2020}.
Even worse, there is no comprehensive throughput analysis of preprocessing pipelines, so one has to estimate the resulting $T_4$ throughput of a pipeline manually for every single deployment to prevent bottlenecks.
Such an estimation is difficult to make, as there are many complexities involved.

One of those is the data encoding after step $S_{\textbf{m}}$, which serializes the entire dataset and places it on a storage device.
The current default way to serialize datasets in two popular DL frameworks is the ``pickle'' encoding for PyTorch~\cite{paszke2019pytorch}, and Protobuf~\cite{protobuf} for TensorFlow~\cite{abadi2016tensorflow}. Both encodings are not optimized for tensor data and may perform poorly. Applying an optimized compression algorithm may be useful but also introduces an additional online decompression step that may affect the $T_4$ throughput.

The deserialization throughput ($T_1$ and $T_3$) depends not only on the encoding but also on the storage solution and its interconnect to the nodes that run the preprocessing pipeline.
A common storage solution in virtualized environments is Ceph~\cite{weil2006ceph}, an object-based storage system.
Such a complex and distributed system's performance depends on the storage hardware and the computing power and must be evaluated on a case-by-case basis.
Without benchmarks for specific hardware setups, it is unclear how to split the pipeline to achieve the maximum throughput~\cite{ra2018understanding,mohan2020analyzing}.

Another variable to consider is the offline preprocessing time, as this may delay the training start.
Long preprocessing times can be prohibitive if not amortized by faster training.

Additionally, some preprocessing steps that feature data augmentation (e.g., random-cropping for images) or shuffling the dataset have to be done online because their results are not deterministic and can not be cached for future epochs.

{\color{diff3}
Some preprocessing steps decrease the dataset size and can make it fit in memory, which would be beneficial over multiple epochs to remove network read effects.
However, preprocessing steps can also be computationally expensive and would better be processed offline which can increase throughput, even if they increase the dataset size.
Both scenarios can benefit the throughput, but it is not obvious to determine without profiling whether caching a dataset or removing a CPU bottleneck is more effective at increasing throughput.
}

In conclusion, deciding how to split a preprocessing pipeline into offline and online steps is a complex problem.
The importance of the trade-offs may depend on individual scenarios, such as preexisting hardware or framework dependencies, which can not be chosen freely.
To solve this problem and provide guidance and best practices to ML engineers and users, a comprehensive analysis of common DL pipelines is needed that provides a structured overview of the pipelines' performance and insights about the individual steps' trade-offs.
It is also necessary to evaluate whether profiling a small sample of the entire dataset is sufficient to estimate the total processing time, storage consumption, and $T_4$ throughput.
This could reduce profiling overhead.
Finally, a software solution should automate the profiling to quickly generate insights for a specific setup to optimize any given pipeline.

\section{Experiments}
\label{sec:experiments}
We analyze four different ML domains to showcase trade-offs in preprocessing pipeline optimizations: CV, NLP, {\color{diff}Audio}, and non-intrusive load monitoring (NILM).
{\color{diff}We evaluate the pipelines with in total seven different datasets in order to compare the impact of different encodings and image resolutions on the respective pipeline's performance.}
Every pipeline is based on common preprocessing steps from popular models and datasets in their respective domains.
We assume that the training throughput is unbounded for our analysis, as we are interested in maximizing $T_4$ irrespective of the actual model.
This section showcases the design of our profiling library, the individual pipelines and defines the experimental setup.

\subsection{PRESTO Library}

After initial manual profiling attempts, we decided to create the \textbf{Pre}processing \textbf{St}rategy \textbf{O}ptimizer (PRESTO) library that automates the generic pipeline profiling process.
The library can be used with any preprocessing pipeline written with the TensorFlow \texttt{tf.data} API~\cite{murray2021tf}, and hence, is readily applicable to different use cases.

PRESTO contains a \texttt{Strategy} wrapper class that splits the preprocessing pipeline at any given step into an offline and online part.
This is done by inserting a serialization and loading step at the \textit{split position} with the TFRecord format, a wrapper around the Protobuf encoding~\cite{protobuf} for TensorFlow.
Additional parameters include the parallelism, sharding, {\color{diff}caching behavior, and compression format}, as well as the temporary logging directory.

The strategy wrapper class executes the entire preprocessing pipeline through the \texttt{tf.data} API.
We simulate the training process by accessing the sample tensor's shape member to measure the preprocessing throughput without training an actual model. 
This allows us to profile the preprocessing pipeline's throughput by calling the \texttt{profile\_strategy()} function.
It accepts two parameters which have to be set manually: \texttt{sample\_count} and \texttt{runs\_total}.
{\color{diff} While it is useful to get an initial understanding of a pipeline's performance with less samples, some bottlenecks only show after local caches are full or a network link is used to its maximum capacity, so that we recommend profiling with the entire dataset.}

Profiling focuses on three key metrics - \textit{preprocessing time}, \textit{storage consumption} and \textit{throughput} - which can be easily tracked by internal Python code.
For more in-depth information, \texttt{dstat} is executed in parallel and provides specific system-level information, like disk read/write loads and network traffic in case of network storage.
These stats and additional metadata, like a unique hash and the split position, are returned as Pandas dataframes~\cite{mckinney-proc-scipy-2010}.

After the profiling is finished, the \texttt{StrategyAnalysis} class summarizes the findings and provides a semi-automatic way to pick the best strategy based on {\color{diff}an objective function.}
The function normalizes the individual metrics to the range of $[0, 1]$ based on their minimum and maximum values and multiplies them by user-defined weights $w_{p, s, t}$.
Let preprocessing time be $\textbf{p}$, storage consumption $\textbf{s}$, and throughput $\textbf{t}$ as vectors of the respective values for all strategies:
\begin{align*}
    f(w_p, w_s, w_t, \textbf{p}, \textbf{s}, \textbf{t}) = w_p \times |\textbf{p}| + w_s \times |\textbf{s}| + w_t \times |\textbf{t}|
\end{align*}
{\color{diff}
The weights $w_{p, s, t}$ are can be defined manually, based on the user's objective.}
As an example, we want to find the optimal strategy to apply hyperparameter tuning on a model before a deadline.
That means we want a low preprocessing time and the highest possible throughput, while the storage consumption is irrelevant. In this case, the weights would look as follows:
\begin{align*}
(w_p, w_s, w_t) = (1,\; 0,\; 1)
\end{align*}
{\color{diff}
On the contrary, if we have access to a cluster with a lot of compute power and are not in a race against time, it will be preferable to sort only by throughput $(w_{p,s} = 0, w_t = 1)$, which is a good default configuration.}
{\color{diff}This procedure can be applied to every strategy, which can have different parallelization, sharding and compression options and lead to new trade-offs.}
{\color{diff}More complex objective functions} can feature cloud providers' processing and storage prices.
We presume that renting a low-cost VM and profiling the different strategies could probe the infrastructure, i.e., network bandwidths.
This allows us to extrapolate the processing performance by tensor-specific CPU benchmarks like PASTA~\cite{li2020parallel} for high-cost VMs.
We provide our library as an open-source project at \href{https://github.com/cirquit/presto}{https://github.com/cirquit/presto}.

\vspace{-0.2cm}
\subsection{Pipelines}

We profiled {\color{diff}seven} pipelines from four different domains and designed them to represent popular DL models.
Table~\ref{tab:experiments:dataset-info} shows the seven datasets we used to profile the pipelines with their storage consumption, sample count, and format.
The datasets and pipelines show a variety of common formats, and different intermediate sizes, e.g., the NLP pipeline has a strategy that increases the initial storage consumption by $64\times$, while NILM has a strategy that decreases the initial storage consumption by a factor of $12\times$ (Fig.~\ref{fig:ss-vs-thr}).
The pipelines were implemented with the \texttt{tf.data} API~\cite{murray2021tf} which automates pipeline execution and allows us to parallelize computations easily.
We define a \textit{sample} in this context as data that is used as input for a DL model.

{\color{diff}The naming of the steps in the pipelines follows a common pattern.
First, the data is read from disk (\texttt{unprocessed}).
After reading the dataset from disk, a concatenation step transforms the input files into a single TFRecord binary in order to allow for efficient sequential read access (\texttt{concatenated}).
The concatenation step was technically not feasible for the Audio pipeline, and was omitted for the NILM pipeline as the raw data was already stored in concatenated binary form.
Then, the data is decoded into a tensor format (\texttt{decoded}).
}
{\color{diff3}
Finally, additional transformation steps can be applied to bring the data into a format suitable for the training process.
Generally, the steps have two characteristics: the online processing time and the relative increase or decrease of storage consumption.
We explain the trade-off between these two characteristics in Sec.~\ref{ssec:storage-versus-throughput}, which can change for the same step just by using different datasets, e.g., decoding can increase or decrease the storage consumption depending on the initial file encoding (e.g., JPG vs. PNG).

The performance of preprocessing steps depends not only on the implementation of the step itself, but also on its position in the pipeline and on the input data. We specifically showcase this behaviour in Sec.~\ref{ssec:modifying-pipeline} by changing the position of a new step in an existing profiled pipeline.
}

\vspace{-0.5cm}
\begin{table}[h]
\scalebox{0.7}{
\begin{tabular}{llrrrl}
\textbf{Dataset} & {\color{diff}\textbf{Pipeline}} & \thead{Sample\\ Count} & \textbf{Size in GB} & \thead{Avg. Sample\\ Size in MB} & \textbf{Format} \\ \hline
ILSVRC2012~\cite{ILSVRC15} & {\color{diff}CV} & 1.3M & 146.90 & 0.1147 & JPG \\ \hline
{\color{diff}Cube++ JPG~\cite{ershov2020cube}} & {\color{diff}CV2-JPG} & 4890 & 2.54 & 0.5203 & JPG \\ \hline
{\color{diff}Cube++ PNG~\cite{ershov2020cube}} & {\color{diff}CV2-PNG} & 4890 & 85.17 & 17.4176 & PNG  \\ \hline
OpenWebText~\cite{Gokaslan2019OpenWeb} & {\color{diff}NLP} & 181K & 7.71 & 0.0427 & TXT \\ \hline
CREAM~\cite{jorde2020cream} & {\color{diff}NILM} & 268K & 39.56 & 0.1477 & HDF5 \\ \hline
Commonvoice (en)~\cite{ardila2019common}  & {\color{diff}MP3} & 13K & 0.25 & 0.0197 & MP3 \\ \hline
Librispeech~\cite{panayotov2015librispeech} & {\color{diff}FLAC} & 29K & 6.61 & 0.2319 & FLAC
\end{tabular}
}
\caption{Metadata of all profiled datasets.}
\label{tab:experiments:dataset-info}
\end{table}
\vspace{-0.8cm}

\subsubsection{CV}

{\color{diff}We profile three datasets with the CV pipeline (Fig.~\ref{fig:cv-pipeline}) to analyze the performance under different image resolutions and encodings. 
ILSVRC2012~\cite{ILSVRC15} is a low resolution, JPG encoded subset of ImageNet~\cite{deng2009imagenet} and is a popular and commonly acknowledged dataset for visual object recognition.
Cube++ is a high-resolution dataset and comes in two flavors: as 16-bit encoded PNGs and JPGs~\cite{ershov2020cube}.
The difference in storage consumption between the two encodings allows for a direct comparison of the decoding performance.
The images from Cube++ are roughly $5\times$ larger than in ILSVRC2012, which allows to analyze how much the image resolution affects the throughput of each strategy.}
Details of this pipeline have been discussed in Section~\ref{sec:introduction}.

\subsubsection{NLP}

\begin{figure}[h]
    \centering
    \begin{subfigure}[b]{0.38\textwidth}
        \centering
        \includegraphics[width=\textwidth]{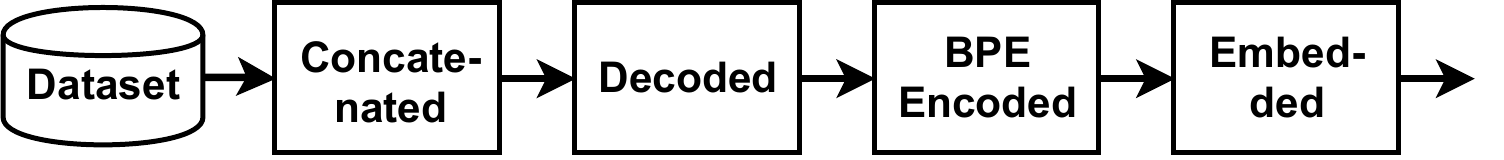}
        \caption[]%
        {{\small NLP }}    
        \label{fig:nlp-pipeline}
    \end{subfigure}
    \hfill
    \begin{subfigure}[b]{0.23\textwidth}  
        \centering 
        \includegraphics[width=\textwidth]{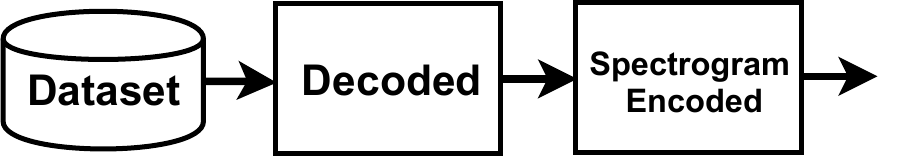}
        \caption[]%
        {{\small MP3 + FLAC}}    
        \label{fig:audio-pipeline}
    \end{subfigure}
    \hfill
    \begin{subfigure}[b]{0.23\textwidth}   
        \centering 
        \includegraphics[width=\textwidth]{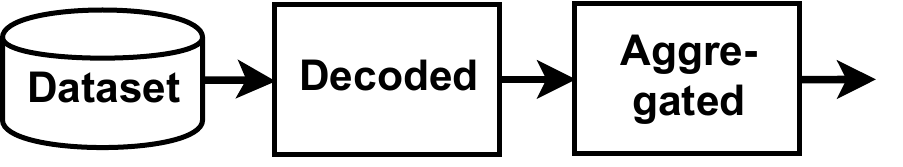}
        \caption[]%
        {{\small NILM }}    
        \label{fig:cream-pipeline}
    \end{subfigure}
    \vspace*{-0.3cm}
    \caption[]
    {\small Preprocessing pipelines} 
    \label{fig:all-pipelines}
\end{figure}

The NLP pipeline  (Fig.~\ref{fig:nlp-pipeline}) is based on GPT-2~\cite{radford2019better}, a popular transformer-based model which tries to predict the next word based on the textual input.
We used the dataset from the corresponding open-source implementation~\cite{Gokaslan2019OpenWeb} of OpenWebText, which was replicated from the GPT-2 paper.
Our version of the dataset is an early iteration and takes up 8\:GB compared to the most current one at 12\:GB.
The dataset consists of HTML content from scraped URLs that have been upvoted on Reddit, a social media platform, as an indicator of human interest and intelligible content. It is stored as multiple text files.

The preprocessing starts with reading text files ({\color{diff}\texttt{concatenated}}) and decoding the actual textual content ({\color{diff}\texttt{decoded}}) with the same HTML parsing library (newspaper~\cite{newspaper2020}) as GPT-2.
Each word is encoded into an \texttt{int32} via Byte Pair Encoding~\cite{DBLP:journals/corr/SennrichHB15} ({\color{diff}\texttt{bpe-encoded}}), which is then looked up in a word2vec embedding~\cite{mikolov2013efficient} that returns a \texttt{float32} tensor of dimension $1\times768$ ({\color{diff}\texttt{embedded}}).
This vector is stacked for every word in the text, resulting in an $n\:\times\:768$ tensor, the final model input.
These preprocessing steps' complexity depends on the tokenization model and the final embedding, making their performance hard to predict.

\vspace{-0.1cm}
\vspace{-0.45cm}

\subsubsection{Audio Processing}

For the audio pipelines, we took inspiration from Baidu's Deep Speech model~\cite{hannun2014deep, amodei2016deep}.
Deep Speech is an RNN-based model that translates spoken audio samples to text.
Both preprocessing pipelines (Fig.~\ref{fig:audio-pipeline}) decode the compressed audio signal into the raw waveform ({\color{diff}\texttt{decoded}}) of the size $l \times r$, where $l$ is the sample duration in seconds and $r$ is the sampling rate encoded as \texttt{int16}.
The waveform is transformed using a short-time Fourier transform (STFT) with a window size of 20\:ms and a stride of 10\:ms.
The spectrogram is then transformed using an 80-bin mel-scale filter bank, leading to a size $\frac{l - 20ms + 10ms}{10ms} \times 80$ tensor encoded as \texttt{float32} ({\color{diff}\texttt{spectrogram-encoded}}).
The difference between the pipelines is their respective input format (MP3 vs. FLAC).
In contrast to some implementations~\cite{hannun2014deep}, we do not convert the data to mel-frequency cepstral coefficients (MFCCs) because it has been found that DL models work as well or better without this transformation~\cite{huzaifah2017comparison, purwins2019deep, solovyev2020deep}.

As datasets, we use the Mozilla Commonvoice 5.1 English corpus~\cite{ardila2019common} for MP3 files and the Librispeech dataset~\cite{panayotov2015librispeech} for FLAC files.

\subsubsection{NILM}

Our signal processing pipeline is based on MEED~\cite{jorde2019meed}, a state-of-the-art event detection model used for non-intrusive load monitoring of electrical data.
The task is to classify individual appliances based on the aggregated voltage and current reading measured on a building's mains.
These datasets typically have a very high frequency, e.g., 6,400-50,000\;Hz~\cite{anderson2012blued, kriechbaumer2018blond, jorde2020cream} to provide information on subtle changes that can be useful for appliance identification.

We used CREAM~\cite{jorde2020cream}, a component-level electrical measurement dataset for two industrial-grade coffeemakers encoded as HDF5 files per hour.
CREAM contains two datasets from two different coffee machines (X8 and X9), from which we used the larger X8 dataset because it takes up more than double the storage consumption of X9, i.e., totals 744 hours of 6.4\:kHz sampled current and voltage.
This dataset's fundamental difference to the other datasets is the \texttt{float64} encoding, which is favorable for NILM tasks~\cite{kahl2017comprehensive} but introduces additional storage consumption.

The pipeline starts by reading HDF5 files and extracts the voltage and current signals from them {\color{diff}(\texttt{decoded})}. They are sliced in 10-second windows, which results in a $2\times64.000$ tensor of \texttt{float64}.
Then, three aggregated values are computed: the reactive power~\cite{barsim2014unsupervised}, the root-mean-square of the current, and its cumulative sum~\cite{jorde2019meed, trung2014event, zhu2018novel} ({\color{diff}\texttt{aggregated}}). 
These aggregation operators work with a dataset period length of 128, which results in a tensor of size $3\times500$ encoded as \texttt{float64}.

\subsection{Experimental Setup}
\label{ssec:experimental-setup}

We execute our experiments on a virtual machine with {\color{diff}80\:GB} DDR4 RAM, 8\:VCPUs on an Intel Xeon E5-2630 v3 8x@2.4\:GHz with an Ubuntu 18.04 image on our OpenStack cluster.
Our Ceph cluster, {\color{diff}backed by HDDs}, is used as a storage device via \textit{cephfs}, with a $10$\:Gb/s uplink and downlink.
This storage is used for both storing the intermediate dataset representations as well as the unprocessed datasets.
We repeat each experiment five times and we drop the page cache after every run to remove memory caching effects {\color{diff}except for explicitly marked caching experiments.
All experiments are run with 8 threads except for explicitly marked scalability experiments with a sharded dataset so that every thread has an assigned individual file to read in parallel.}
All experiments are executed with Python 3.7 and TensorFlow 2.4.
Specific library versions are available in our GitHub repository.
\section{Analysis}
\label{sec:analysis}
By profiling the preprocessing pipelines, we aim to provide insights about how to pick the optimal preprocessing strategy for a specific set of hardware, the datasets, and the characteristics of each pipeline.
The goal is to maximize the $T_4$ throughput (Fig.~\ref{fig:generic-dl-pipeline}) while keeping the storage consumption and offline preprocessing time low.
Our analysis is focused on four core aspects:
\begin{description}[align=left]
    \item[Storage Consumption versus Throughput:] 

    {\color{diff}A high storage consumption can render extensive offline preprocessing useless or even counterproductive to achieve high throughput.
    This has two causes. First, the dataset may be split into many files, and the storage does not respond fast enough to random read access, leading to a storage bottleneck.
    Secondly, specific preprocessing steps (e.g., normalizing an integer range to floating points) inflate the data volume.
    This can lead to a lower throughput because the saved preprocessing time is outweighed by the increased data ingestion time (see Section~\ref{ssec:storage-versus-throughput}).}

    \item[Caching:]

    {\color{diff}As a DL training process typically iterates over the dataset multiple times, there can be an increased throughput due to caching effects after the first epoch.
    However, this effect depends on whether the preprocessed training data fits into memory.
    Further, reading the cached dataset from memory can help to isolate processing bottlenecks from storage bottlenecks (see Section~\ref{ssec:caching}).}

    \item[Compression:]

    {\color{diff}Compression provides a way to trade off CPU overhead for en-/decode steps against smaller storage consumption.
    We profile each strategy with GZIP~\cite{10.17487/RFC1952} and ZLIB~\cite{10.17487/RFC1950} compression and show under which circumstances compression improves the throughput (see Section~\ref{ssec:compression}).}



    \item[Parallelization Capabilities:]
    Preprocessing steps can hinder performance if multi-core CPUs are not utilized effectively.
    Scalability bottlenecks may have a substantial impact on throughput.
    We compare the speedup of each preprocessing step under multi-threading and {\color{diff}system-level caching} in Section~\ref{ssec:parallelization-capabilities}.

\end{description}
We also touch on shuffling (see Section~\ref{ssec:shuffling}) {\color{diff}and discuss how to introduce new steps to an already profiled CV pipeline based on a case study in Section~\ref{ssec:modifying-pipeline}.}
\vspace{-0.25cm}
\begin{table}[h]
\scalebox{0.8}{
\begin{tabular}{r|r|r|r|r}
\textbf{Threads} & \textbf{Files per Thread} & \textbf{Bandwidth} & \textbf{Latency} & \textbf{IOPS} \\ \hline
 1 & 1 & 219\:MB/s & $4-10\mu\text{s}$ &  53400 \\
 8 & 1 & \textbf{910\:MB/s} & $4-10\mu\text{s}$ & \textbf{222000} \\ \hline
 1 & 5000 & 6.6\:MB/s & $4-10\mu\text{s}$ & 1629 \\
 8 & 5000 & 40.4\:MB/s & $4-10\mu\text{s}$ & 9853 
\end{tabular}
}
\caption{{\color{diff}\texttt{fio} profile of our storage cluster}}
\label{tab:storage-bw}
\end{table}
\vspace{-0.9cm}

\subsection{Storage Consumption versus Throughput}
\label{ssec:storage-versus-throughput}

We profiled the throughput and storage consumption for all strategies of each pipeline in Figure~\ref{fig:ss-vs-thr}.
{\color{diff}The theoretical network read speed to the HDD-backed Ceph storage cluster is capped at 1.25\:GB/s (10\:Gb/s) due to hardware limitations, but the actual rates differ based on the access pattern.
To provide a pipeline-independent measurement, we profiled four workloads with the \texttt{fio} tool~\cite{fio} to simulate both sequential and random file access with 5000 files of 0.2\:MB each and with one 5\:GB file, which is comparable to our \texttt{unprocessed} and \texttt{concatenated} strategies.
Additionally, we tested the single-threaded performance compared to 8 threads with the same workload per thread.
Table~\ref{tab:storage-bw} shows that reading sequentially is 33$\times$ faster for single- and 22$\times$ faster for multi-threaded execution.
While our single-threaded bandwidth is limited to 219\:MB/s, the multi-threaded execution is close to the hardware cap with 910\:MB/s.
This helps to explain our main observations:}

{\color{diff}
\vspace{-0.2cm}
\begin{table}[h]
\scalebox{0.8}{
\begin{tabular}{l|r|r|r|r}
\textbf{Pipeline} & \multicolumn{2}{c|}{\textbf{Throughput in SPS}} & \multicolumn{2}{c}{\textbf{Network Reads in MB/s}}  \\ \hline
                  & \texttt{unprocessed} & \texttt{concatenated} & \texttt{unprocessed} & \texttt{concatenated}  \\ \hline
CV                & 107 & \textbf{962} & ($\pm 3$) 12 & ($\pm 16$) \textbf{111} \\ \hline
CV (SSD)          & 588 & \textbf{944} & ($\pm 8$) 68 &    ($\pm 15$) \textbf{108} \\ \hline
CV2-JPG           & 88  & \textbf{288} & ($\pm 11$) 46 &    ($\pm 72$) \textbf{110}  \\ \hline
CV2-PNG           & 15  & \textbf{21}  & ($\pm 37$) 270 &   ($\pm 54$) \textbf{390}  \\ \hline
NLP               & 6   & 6   & ($\pm 0.2$) 0.21 &  ($\pm 1.5$) \textbf{0.26}  \\ \hline
NLP (SSD)         & 3   & 3   & ($\pm 0.2$) \textbf{0.17} &  ($\pm 1.2$) 0.16  \\
\end{tabular}
}
\caption{{\color{diff}Throughput and average network read speeds for strategies with concatenation.}}
\label{tab:concatenated-throughput-reads}
\end{table}
\vspace{-0.7cm}

\textbf{(1) Concatenating can increase throughput significantly.}

An I/O bottleneck may arise when the storage cannot saturate the hardware bandwidth based on the data access pattern.
Out of the four pipelines which have a \texttt{concatenated} strategy (CV, CV2-JPG, CV2-PNG, and NLP), we see that all CV-based pipelines have a throughput increase between 1.4$\times$ and 9$\times$ compared to the \texttt{unprocessed} strategy (Table~\ref{tab:concatenated-throughput-reads}).
The individual differences in the throughput can be explained by the dataset size and the average storage consumption of a sample (Table~\ref{tab:experiments:dataset-info}).
Due to the CV samples being smaller than the CV2-JPG samples, CV achieves 962\:SPS compared to 288\:SPS while having a similar network read speed of approximately 110\:MB/s.
The CV2-PNG dataset has a sample storage consumption of 17.4\:MB and the network read speed increases from 270\:MB/s to 390\:MB/s with concatenation.

Contrary to CV-based pipelines, the NLP pipeline does not benefit from concatenating as the throughput stays at 6\:SPS, indicating a CPU bottleneck.
This bottleneck is resolved in the \texttt{decoded} strategy, where throughput increases significantly (Fig.~\ref{fig:ss-vs-thr-nlp}).

As an HDD-based storage solution is particularly vulnerable to random access bottlenecks, we additionally profiled the performance of the CV and NLP pipeline on an SSD-backed Ceph cluster.
The CV \texttt{unprocessed} strategy had a throughput of 588\:SPS, which is almost $6\times$ faster than the HDD experiments.
However, at the \texttt{concatenated} strategy, both HDD and SSD reach roughly the same throughput (962\:SPS vs. 944\:SPS), i.e., at sequential access the SSD-backed storage is not faster.
For NLP, the SSD storage did not provide a better throughput as it still faces the CPU bottleneck at the \texttt{concatenated} strategy.

\vspace{-0.1cm}
\begin{figure}[h]
    \begin{subfigure}[c]{0.22\textwidth}
        \includegraphics[width=\textwidth]{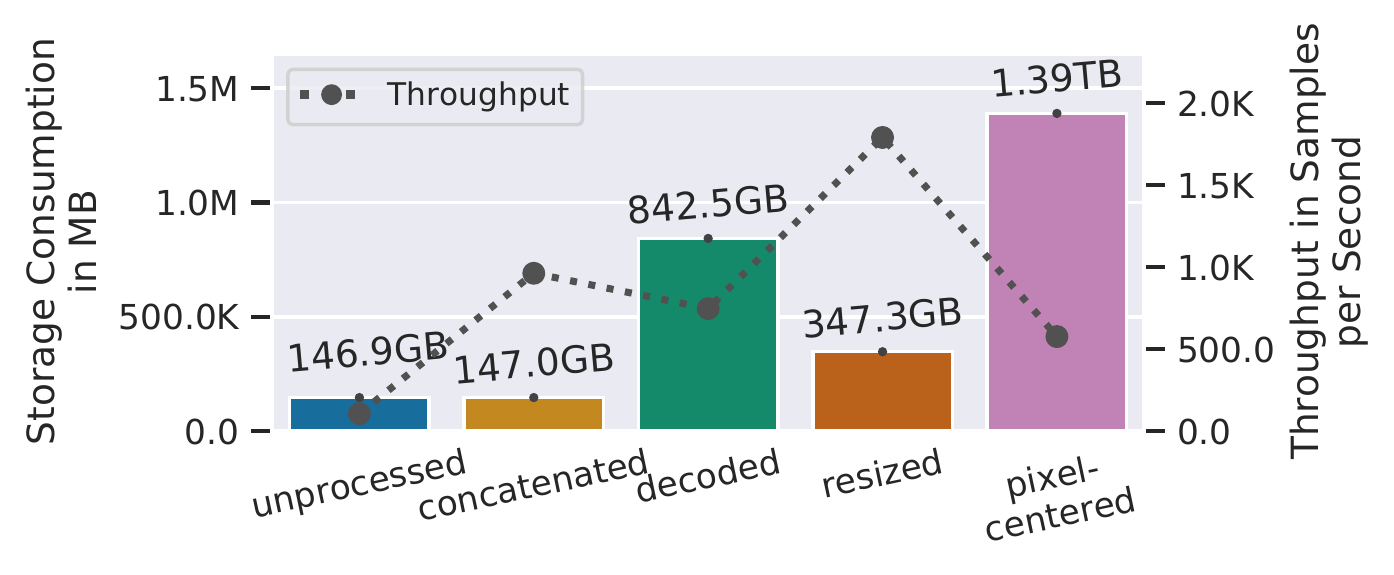}
        \vspace{-18pt}
        \caption{CV}
        \label{fig:ss-vs-thr-cv}
    \end{subfigure}
    \begin{subfigure}[c]{0.22\textwidth}
        \includegraphics[width=\textwidth]{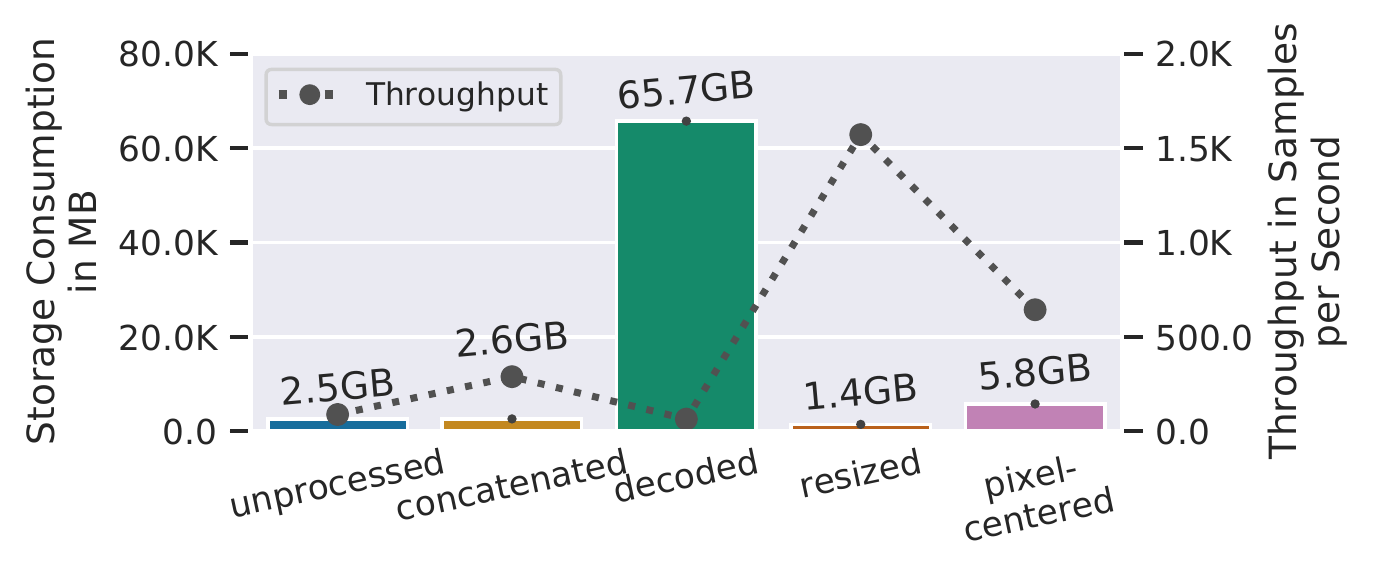}
        \vspace{-18pt}
        \caption{CV2-JPG}
        \label{fig:ss-vs-thr-cv2-jpg}
    \end{subfigure}
    \begin{subfigure}[c]{0.22\textwidth}
        \includegraphics[width=\textwidth]{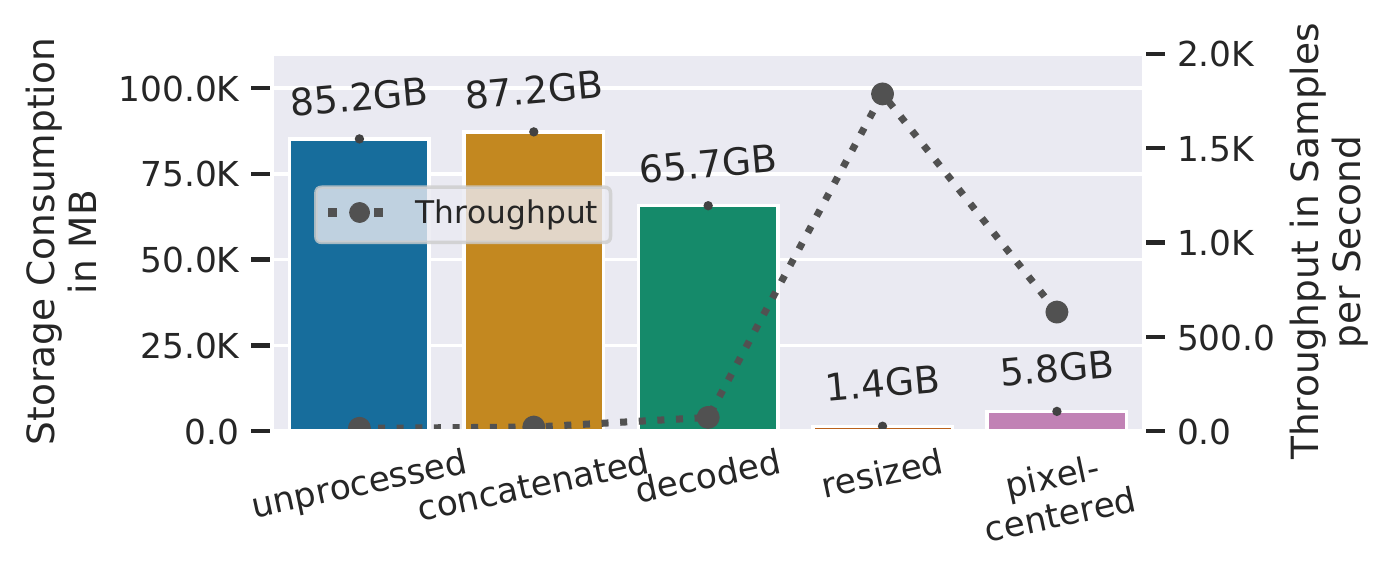}
        \vspace{-18pt}
        \caption{CV2-PNG}
        \label{fig:ss-vs-thr-cv2-png}
    \end{subfigure}
    \begin{subfigure}[c]{0.22\textwidth}
        \includegraphics[width=\textwidth]{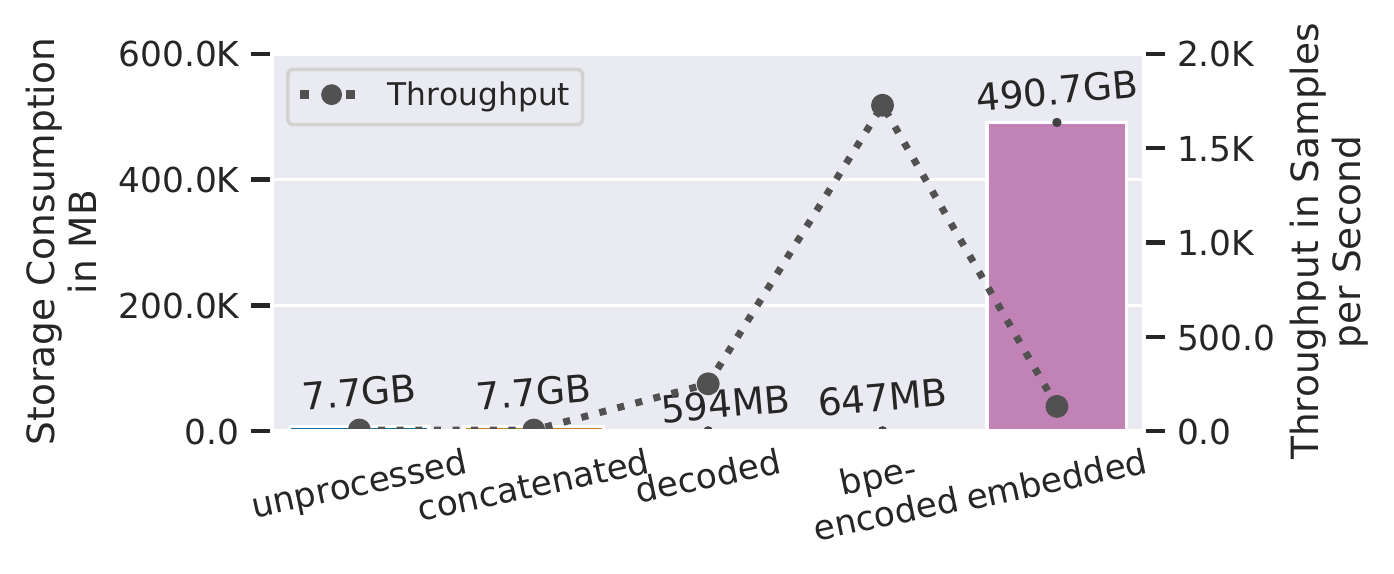}
        \vspace{-18pt}
        \caption{NLP}
        \label{fig:ss-vs-thr-nlp}
    \end{subfigure}
    \begin{subfigure}[c]{0.22\textwidth}
        \includegraphics[width=\textwidth]{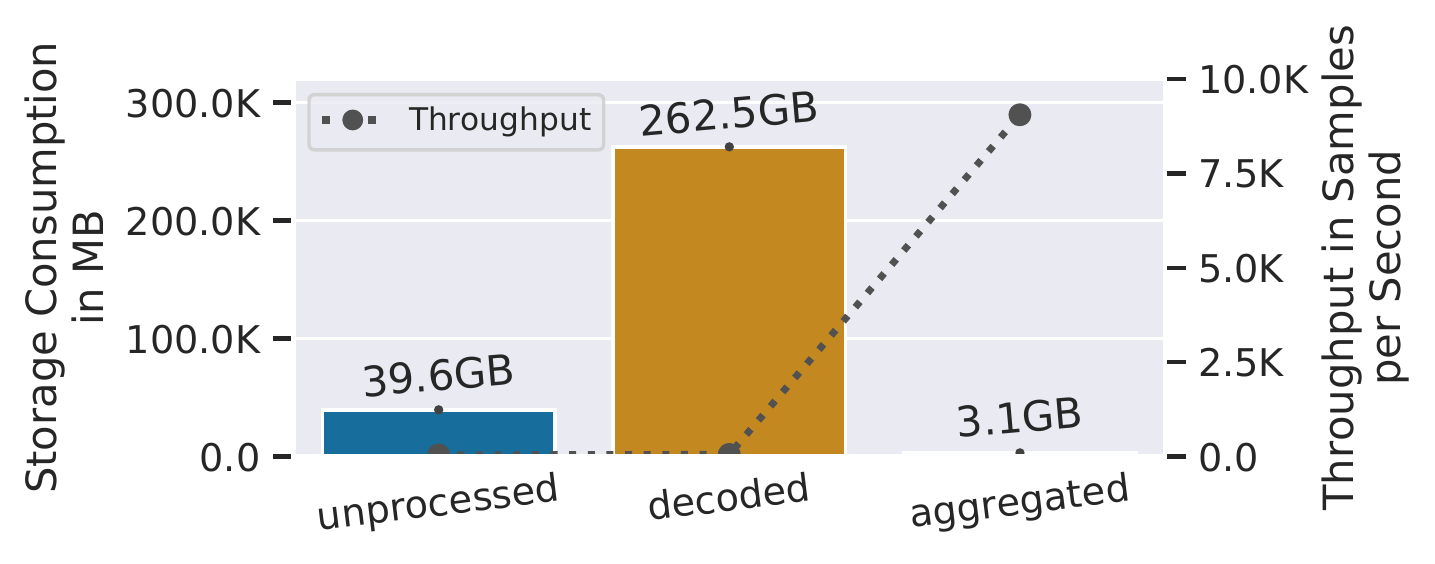}
        \vspace{-18pt}
        \caption{NILM}
        \label{fig:ss-vs-thr-nilm}
    \end{subfigure}
    \begin{subfigure}[c]{0.22\textwidth}
        \includegraphics[width=\textwidth]{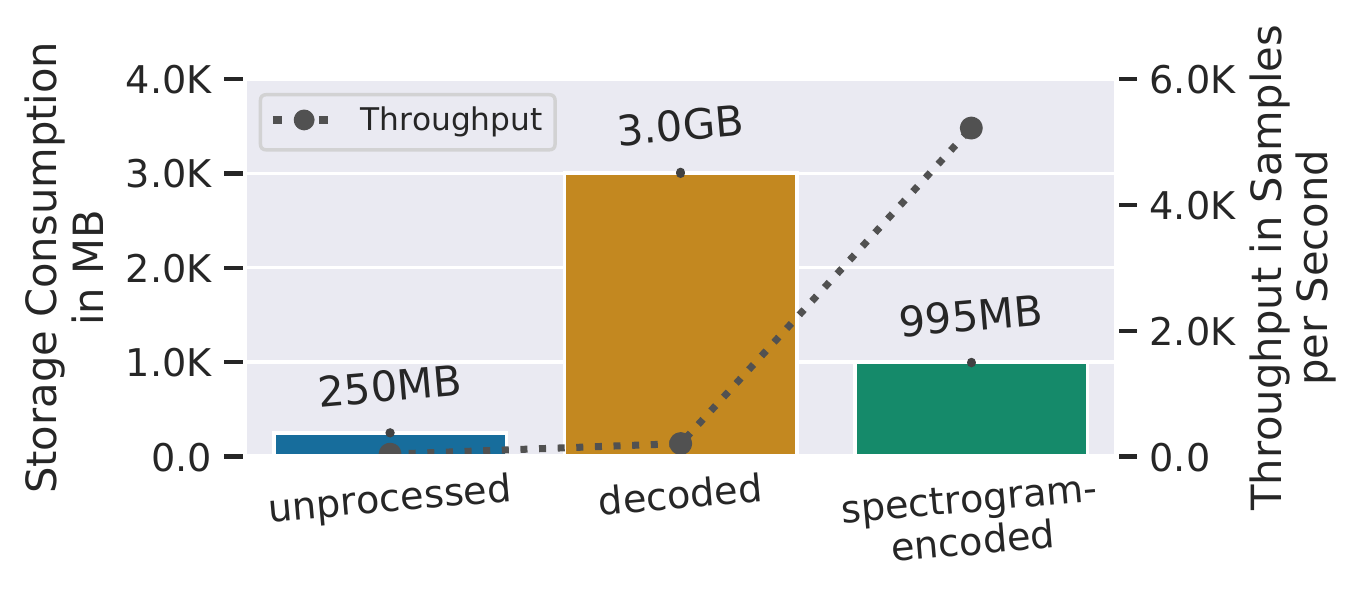}
        \vspace{-18pt}
        \caption{MP3}
        \label{fig:ss-vs-thr-mp3}
    \end{subfigure}
    \begin{minipage}[c]{0.22\textwidth}
        \begin{subfigure}[c]{\textwidth}
            \includegraphics[width=\textwidth]{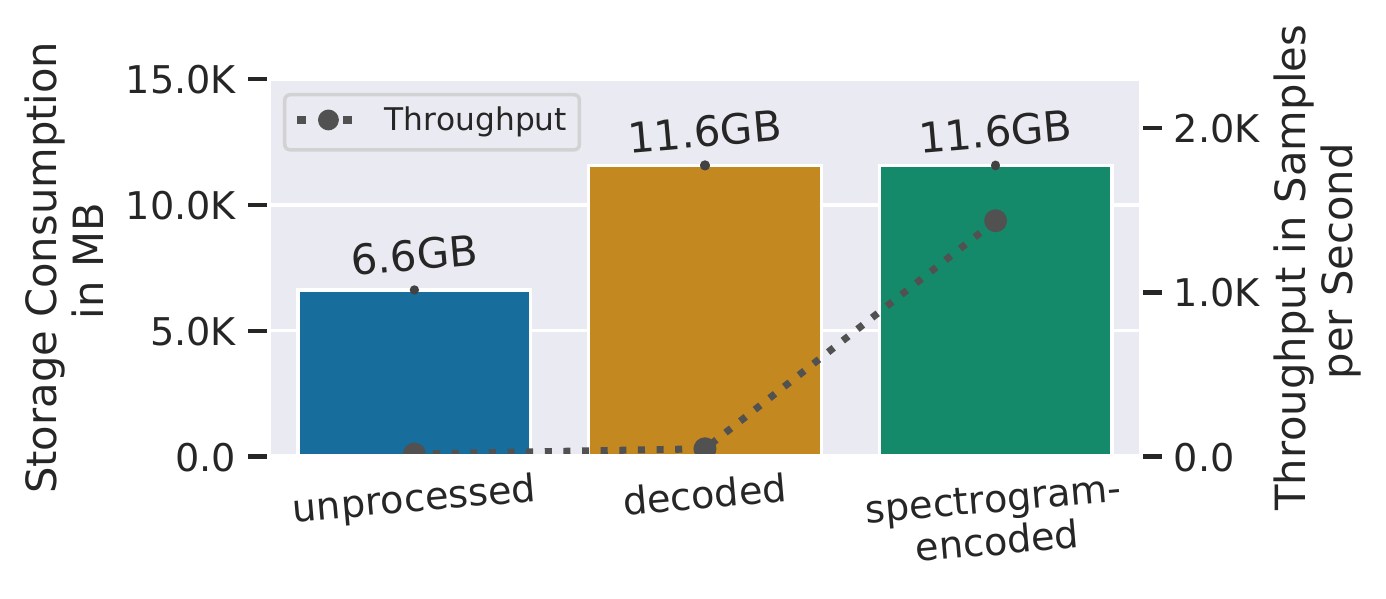}
            \vspace{-18pt}
            \caption{FLAC}
            \label{fig:ss-vs-thr-flac}
        \end{subfigure}
    \end{minipage}\hspace{5mm}
    \begin{minipage}[c]{0.20\textwidth}
        \vspace{-10pt}
        \caption{Storage consumption (left y-axis) for each dataset representation compared to the $T_4$ throughput (dotted line, right y-axis).}
        \label{fig:ss-vs-thr}
    \end{minipage}
\end{figure}

\textbf{(2) The maximum throughput of a strategy is influenced by its storage consumption. }

When the CPU performance in combination with a storage setup can saturate the hardware bandwidth (i.e., in our case, read data with 1.25\:GB/s), then a maximum theoretical throughput can be calculated by dividing the \textit{bandwidth} by the \textit{storage consumption per sample}.
This theoretical throughput is based on two steps.
First, one reads a sample from the storage into memory.
Second, one applies the online transformation steps in succession until the sample can be fed into the training process.
The total time of these two steps defines the pipeline's throughput (i.e., samples per second), hence also the network read speed (i.e., MB per second).
In our case, the actual throughput we can achieve is bound by the multi-threaded read performance to our cluster, which is at 910MB/s with eight threads (Table~\ref{tab:storage-bw}).
This profiled network read speed provides a baseline of the maximum possible throughput.
The goal of every strategy should be to have a short enough transformation step to achieve this baseline read speed.
In turn, if transformation steps are too long, such that the maximum read cannot be reached, we can assume a CPU bottleneck.

A characteristic result of how storage consumption affects the throughput of strategies can be seen in the CV, CV2-JPG, CV2-PNG and NLP pipelines (Fig.~\ref{fig:ss-vs-thr-cv}, \ref{fig:ss-vs-thr-cv2-jpg}, \ref{fig:ss-vs-thr-cv2-png}, \ref{fig:ss-vs-thr-nlp}), where the last strategy has the least amount of online processing to do, but performs worse than its corresponding preceding strategy.
An excellent example of this is the CV pipeline.
At the last strategy, \texttt{pixel-centered}, we have an average network read speed of 585\:MB/s and need to read 1.4\:TB of data.
This stands in contrast to the previous strategy, \texttt{resized}, which has a lower network read speed of 470\:MB/s, but only needs to read 347\:GB.
Therefore, the \texttt{resized} strategy has a more than $3\times$ greater throughput of 1789\:SPS compared to \texttt{pixel-centered} (576\:SPS), even though \texttt{resized} applies more processing steps online.
The cause of the increased storage consumption is that \texttt{pixel-centered} converts each pixel from an \texttt{uint8} to a \texttt{float32} which effectively quadruples the storage consumption.
All our CV-based pipelines share this characteristic at different magnitudes, which results in the \texttt{resized} strategy having the best throughput.

}

{\color{diff2}
The NLP pipeline has a similar issue with the embedded step, which slows down the throughput from 1726\:SPS with \texttt{bpe-encoded} to 131\:SPS with \texttt{embedded} (a factor of $13\times$).
Applying the embedding step online is very computationally intensive, which yields a data ingestion of only 6\:MB/s for \texttt{bpe-encoded}.
One could think that preprocessing this step offline should improve performance.
But this is not the case, because the storage consumption increases from 647\:MB to 491\:GB, such that the benefit of processing the embedding step offline is outweighed by the increased time to read the dataset.
}

{\color{diff}
The remaining pipelines, NILM, MP3 and FLAC (Fig.~\ref{fig:ss-vs-thr-nilm},~\ref{fig:ss-vs-thr-mp3},~\ref{fig:ss-vs-thr-flac}), share the common characteristic that the last preprocessing step is the most computationally expensive one, which leads to the best throughput when processed offline.
While they all have different storage consumption, none of the pipelines approaches the maximum possible network read speeds at their respective last strategy.

On first sight, this is counter intuitive.
In the last strategy, there is almost no processing to be done except for decoding the read data.
Why do these strategies not approach the network read limit?
A deeper investigation leads to our following observation.

\vspace{-0.44cm}
\begin{figure}[h]
    \centering
    \includegraphics[width=0.47\textwidth]{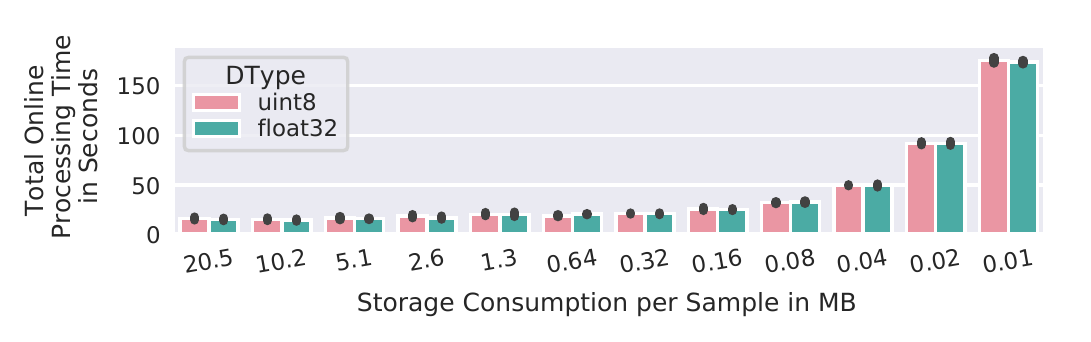}
    \vspace{-14pt}
    \caption{{\color{diff}Profiling a synthetic 15\:GB dataset with different datatypes and sample sizes.}}
    \label{fig:synthetic-dataset}
\end{figure}
\vspace{-0.3cm}

\textbf{(3) A high storage consumption per sample allows for easier I/O bandwidth saturation.}

A deserialization step is applied onto every sample which is read from the storage to transform it from a bytestream into a tensor.
We observed that an increasing sample size positively influences the I/O bandwidth of reading and deserializing.
To provide a basis to our observation, we conduct an experiment with synthetic data that shows the effect of different sample sizes on the online processing time of reading and deserializing.
We evaluate sample sizes from 0.01\:MB to 20.5\:MB with doubling increments while keeping a total storage consumption of 15\:GB for both \texttt{uint8} and \texttt{float32} datatypes which are common in our real-world pipelines.
}
{\color{diff3}
To keep the same total storage consumption with different sample sizes we adapted the sample count, which ranged from 732 (20.5\:MB) to 1.5M (0.01\:MB) samples.
}
{\color{diff}
Figure~\ref{fig:synthetic-dataset} shows the result that reading the same amount of data with different sample sizes has a major effect on the processing time.
A dataset consisting of large (20.5\:MB) samples takes less than half the processing time of small ($\leq0.08$\:MB) samples.
At a sample size of 0.01\:MB, it takes more than $11\times$ longer to process the 15\:GB of data compared to 20.5\:MB samples.
Finally, the different data types do not have an impact on the processing time, as both \texttt{uint8} and \texttt{float32} show similar results.

A good example for this observation is the comparison of the \texttt{decoded} strategy between CV (Fig.~\ref{fig:ss-vs-thr-cv}) and CV2-JPG (Fig.~\ref{fig:ss-vs-thr-cv2-jpg}).
The average sample size is 13\:MB for the \texttt{decoded} strategy of CV2-JPG with a network read speed of 828\:MB/s, which indicates an I/O bottleneck.
However, with the same strategy, the CV pipeline has a sample size of 0.6\:MB and the average network read rate is at 491\:MB/s, which is not even close to our maximum bandwidth.
Notably, the CV pipeline has a lower computational load compared to CV2-JPG \textit{and} has to read fewer data from storage with each sample due to the small sample size.
But it still does not manage to saturate the I/O bandwidth.
Therefore, the CV \texttt{decoded} strategy suffers from a CPU bottleneck.
To further validate our assumption, we profiled the CV pipeline with 16 threads which increased the network read speed by 64\:MB/s and improved the throughput by 142\:SPS.
The additional multi-threading also increased the throughput by 583\:SPS and 100\:SPS for the \texttt{resized} and \texttt{pixel-centered}, respectively.
All last strategies like \texttt{aggregated} (Fig.~\ref{fig:ss-vs-thr-nilm}), \texttt{spectrogram-encoded} (Fig.~\ref{fig:ss-vs-thr-mp3},~\ref{fig:ss-vs-thr-flac}), and \texttt{embedded} (Fig.~\ref{fig:ss-vs-thr-nlp}) share that characteristic and do not saturate the I/O bandwidth (96\:MB/s, 317\:MB/s, 564\:MB/s and 315\:MB/s respectively).

}

{\color{diff}
\vspace{-0.3cm}
\subsection{Caching}
\label{ssec:caching}

DL training jobs typically run over multiple epochs, which means the dataset is read multiple times and could benefit from being cached in memory after the first epoch.
We evaluated the throughput of all pipelines over two epochs for all strategies.
In this set of experiments, we do not flush the page cache after the first epoch.
Our observations are as follows:

\textbf{(1) Caching is not beneficial when the storage consumption is higher than the available memory.}

If the data set is too large for the memory, the dataset is read completely from the storage at every epoch.
Hence, throughput is not increased by caching.
All strategies that have a storage consumption higher than 80\:GB (Fig.~\ref{fig:ss-vs-thr}) have the same throughput over all epochs (Fig.~\ref{fig:caching}).

\textbf{(2) Caching does not remove CPU bottlenecks. }

Assuming that the dataset fits into memory, caching can only improve throughput significantly if there is no CPU bottleneck.
Reading data from memory is much faster than from remote network storage, but the impact of fast data access can become insignificant when followed by computationally expensive preprocessing steps.
An excellent example of this is the NLP pipeline (Fig.~\ref{fig:caching-nlp}).
The first two strategies \texttt{unprocessed} and \texttt{concatenated} have the same throughput of 6\:SPS over all epochs because decoding is very compute intensive while the datasize is relatively small (7.7\:GB).
Then, after decoding the data (594\:MB), we face a new computationally expensive step, byte-pair encoding, which transforms the text into integers and increases the storage consumption to 647\:MB.
This is followed by the embedding step, which also takes much time.
All of these strategies face a CPU bottleneck while having a storage consumption that allows the data to be cached.
Finally, at the \texttt{embedded} strategy, the dataset only needs to be read from the storage, but grows in size to 490.7\:GB, such that caching has no impact on throughput.

Similar CPU bottlenecks can also be observed in the \texttt{unprocessed} strategies of CV2-PNG, NILM, MP3 and FLAC (Fig.~\ref{fig:caching-cv2-png},~\ref{fig:caching-nilm},~\ref{fig:caching-mp3},~\ref{fig:caching-flac}), the \texttt{concatenated} strategies of CV2-\{JPG,PNG\} (Fig.~\ref{fig:caching-cv2-jpg},~\ref{fig:caching-cv2-png}) and the \texttt{decoded} strategies of MP3 and FLAC.
The remaining strategies (\texttt{resized}, \texttt{pixel-centered}, \texttt{aggregated}, \texttt{spectrogram-encoded}) benefit from caching the most as they have low storage consumption and are not followed by computationally expensive steps.
However, caching improves the throughput with differing factors ($1.1\times$-$4.2\times$), which results in the next observation.

\begin{figure}[h]
    \begin{subfigure}[c]{0.22\textwidth}
        \includegraphics[width=\textwidth]{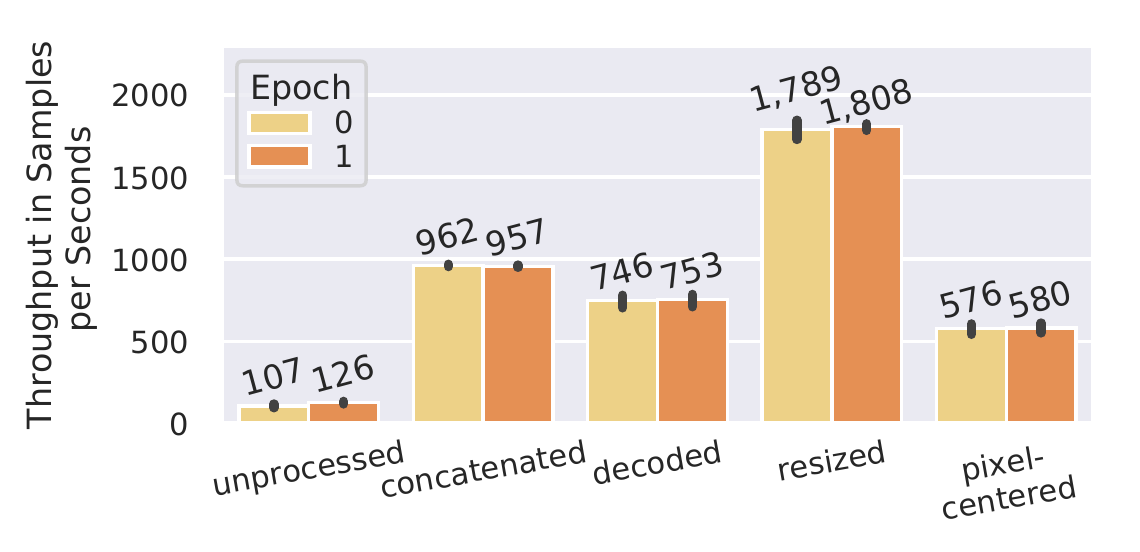}
        \vspace{-18pt}
        \caption{CV}
        \label{fig:caching-cv}
    \end{subfigure}
    \begin{subfigure}[c]{0.22\textwidth}
        \includegraphics[width=\textwidth]{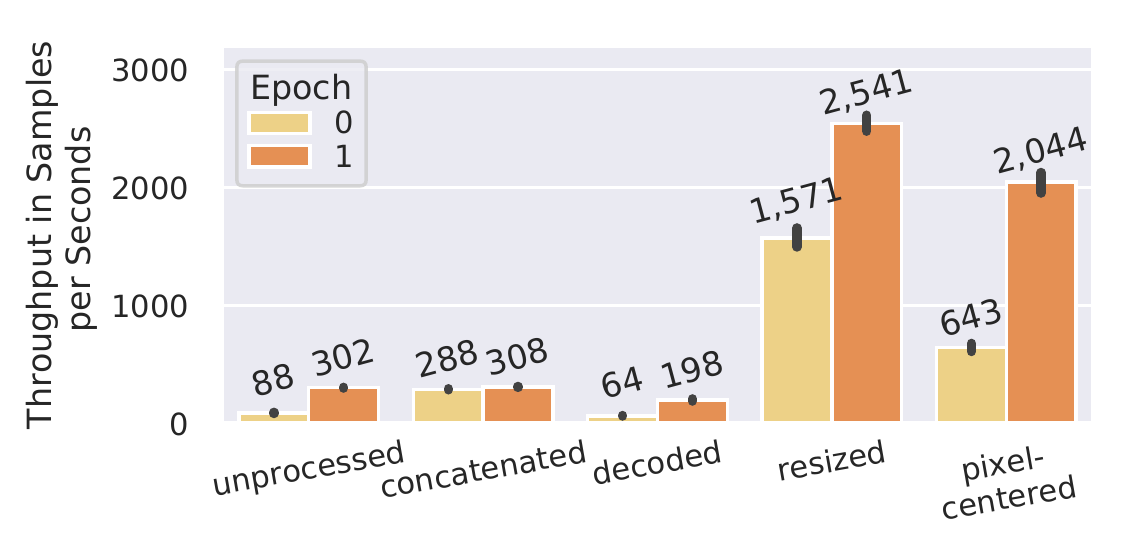}
        \vspace{-18pt}
        \caption{CV2-JPG}
        \label{fig:caching-cv2-jpg}
    \end{subfigure}
    \begin{subfigure}[c]{0.22\textwidth}
        \includegraphics[width=\textwidth]{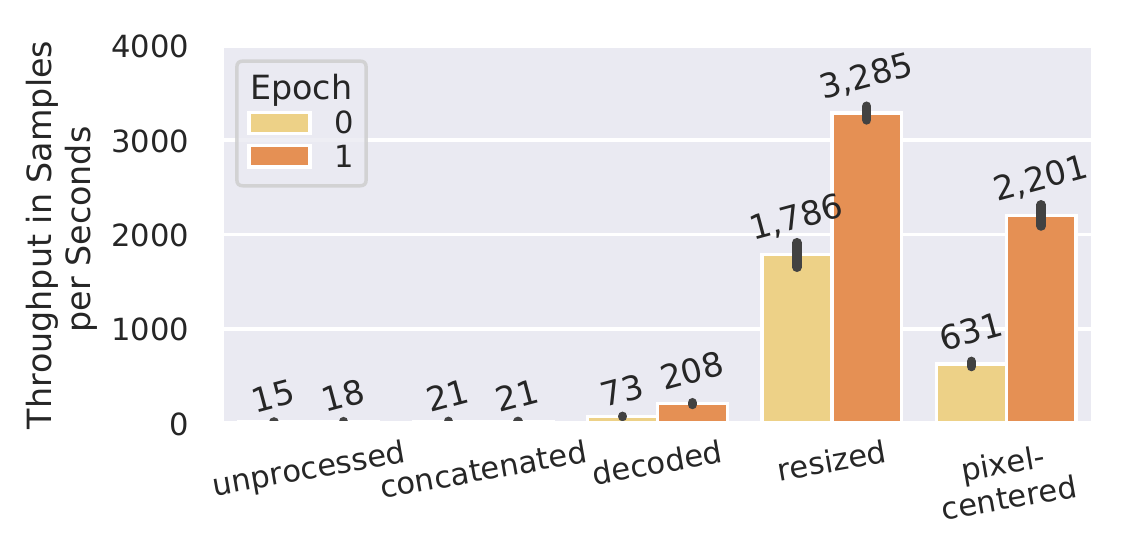}
        \vspace{-18pt}
        \caption{CV2-PNG}
        \label{fig:caching-cv2-png}
    \end{subfigure}
    \begin{subfigure}[c]{0.22\textwidth}
        \includegraphics[width=\textwidth]{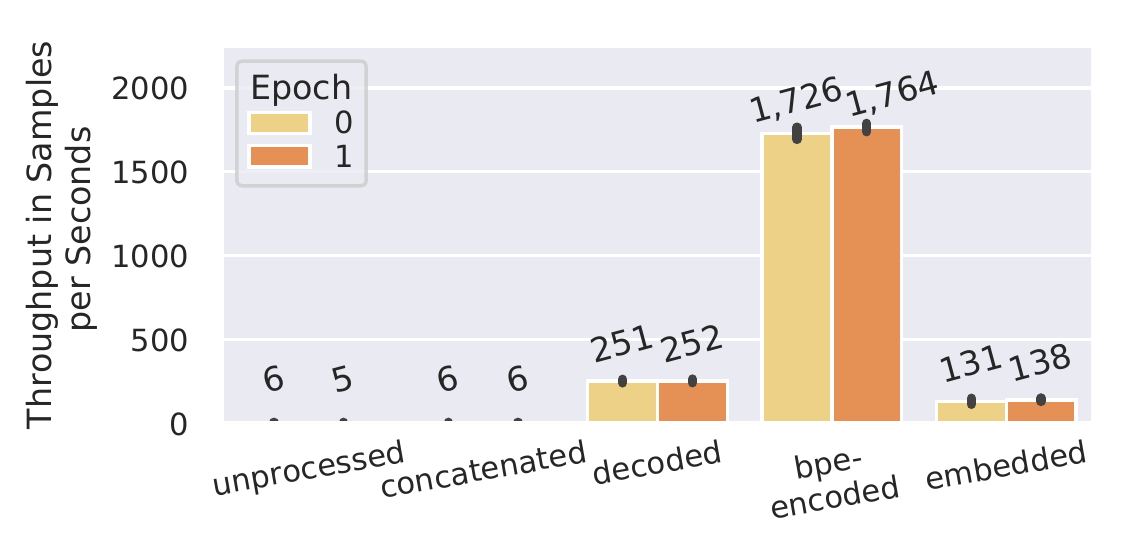}
        \vspace{-18pt}
        \caption{NLP}
        \label{fig:caching-nlp}
    \end{subfigure}
    \begin{subfigure}[c]{0.22\textwidth}
        \includegraphics[width=\textwidth]{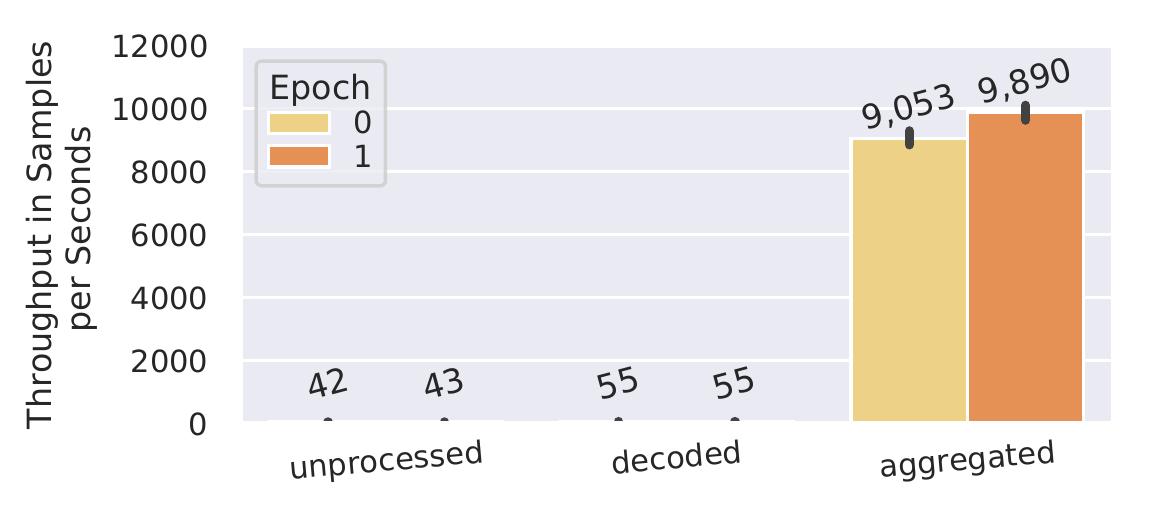}
        \vspace{-18pt}
        \caption{NILM}
        \label{fig:caching-nilm}
    \end{subfigure}
    \begin{subfigure}[c]{0.22\textwidth}
        \includegraphics[width=\textwidth]{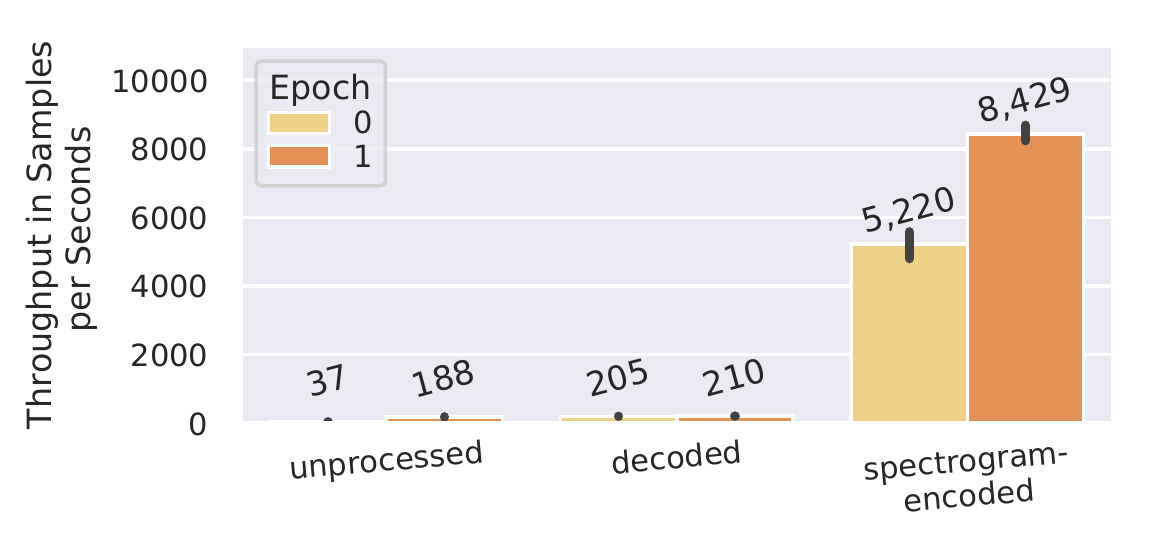}
        \vspace{-18pt}
        \caption{MP3}
        \label{fig:caching-mp3}
    \end{subfigure}
    \begin{minipage}[c]{0.22\textwidth}
        \begin{subfigure}[c]{\textwidth}
            \includegraphics[width=\textwidth]{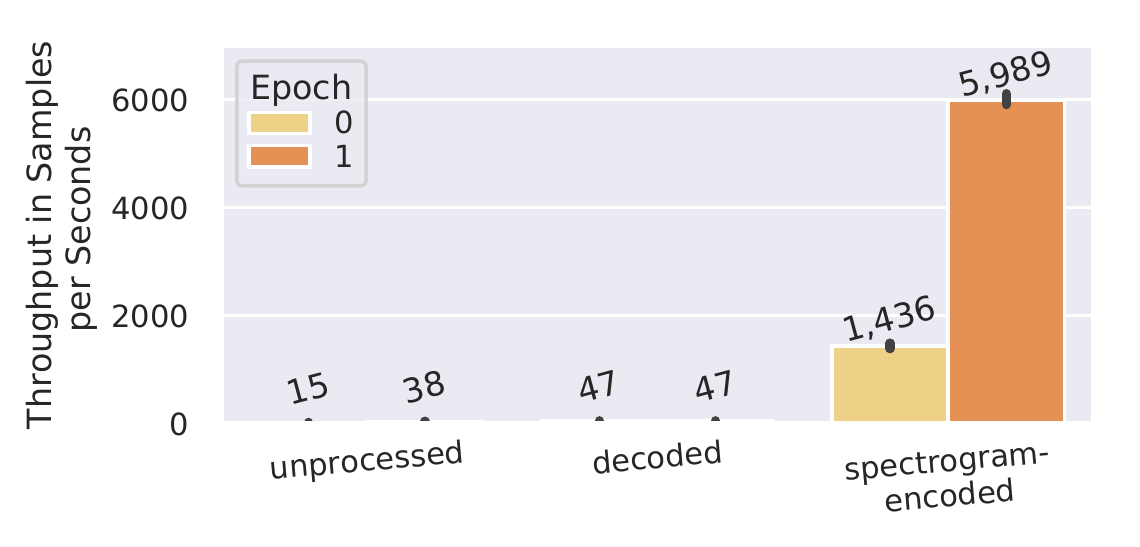}
            \vspace{-18pt}
            \caption{FLAC}
            \label{fig:caching-flac}
        \end{subfigure}
    \end{minipage}\hspace{5mm}
    \begin{minipage}[c]{0.20\textwidth}
        \vspace{-18pt}
        \caption{{\color{diff}Effects of caching on $T_4$ throughput for all pipelines.}}
        \label{fig:caching}
    \end{minipage}
\end{figure}
\vspace{-0.3cm}

\textbf{(3) System-level caching performance is affected by the storage consumption per sample. }

While cached data removes the performance impact of remote storage, the preprocessed dataset still has to be fetched from memory and deserialized.
We confirmed this by comparing the trace log between epochs.
To examine the memory bandwidth, we used \texttt{sysbench}~\cite{sysbench} to profile our memory which resulted in 166\:GB/s.
This should theoretically yield a multiplicative increase in throughput, which we do not achieve because we are not close to the maximum I/O bandwidth (cf. Sec.~\ref{ssec:storage-versus-throughput} observation \textbf{(3)}).

\begin{figure}[h]
    \includegraphics[width=0.47\textwidth]{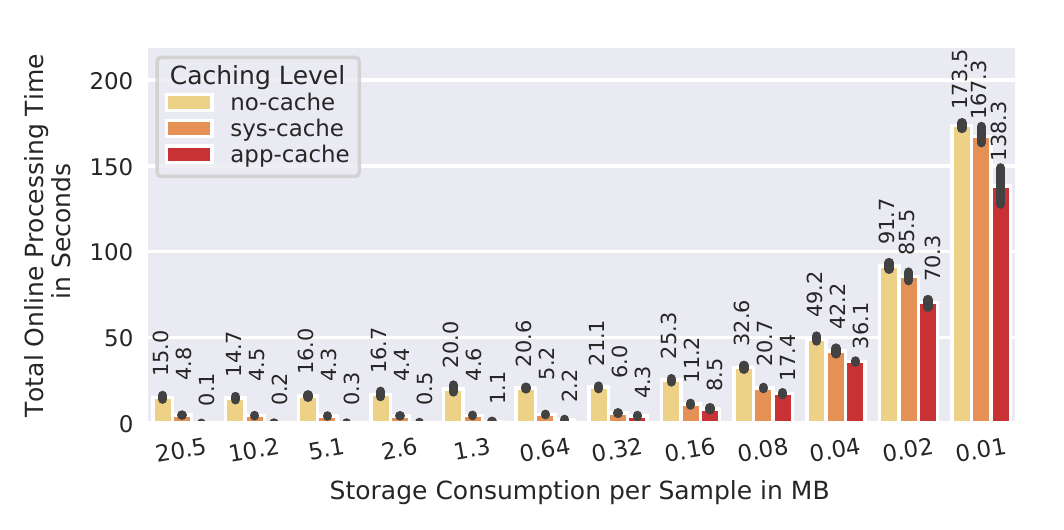}
    \vspace{-14pt}
    \caption{{\color{diff}Online processing time for different caching levels and sample sizes of a synthetic 15\:GB dataset.}}
    \label{fig:synthetic-dataset-cached-general}
\end{figure}
}

{\color{diff2}

We investigate this observation by profiling synthetic \texttt{float32} datasets with different sample sizes with both system- and application-level caching (Fig.~\ref{fig:synthetic-dataset-cached-general}).
At the lower end of storage consumption, starting with 0.16\:MB per sample, reading smaller samples takes increasingly longer.
The smaller the sample size, the more processing time the deserialization takes, which lessens the final throughput of the cached dataset.
At 0.04\:MB and lower, the processing time when data is cached in memory (\texttt{sys-cache}) is comparable to the case where data is on storage (\texttt{no-cache}), nullifying the effects of caching.

}

{\color{diff}

Our real-world experiments show the same behaviour.
The storage consumption per sample for MP3 is 0.08\:MB at the \texttt{spectrogram\\-encoded} strategy, while having a relative throughput increase of $1.6\times$ (Fig.~\ref{fig:caching-mp3}) with caching.
Meanwhile, the FLAC pipeline has a storage consumption of 0.4\:MB per sample with the same strategy and increases its throughput by $4.2\times$ (Fig.~\ref{fig:caching-flac}) with caching.
The NILM pipeline shows almost no increase in throughput ($1.1\times$) over multiple epochs (Fig.~\ref{fig:caching-nilm}) as the sample size is only 0.012\:MB.

While this is interesting, system-level caching via the page cache is somewhat unsatisfactory, since one wants to cache the tensor data and not be bottlenecked by the deserialization.
The TensorFlow function \texttt{tf.data.Dataset.cache} caches the deserialized tensors in memory, avoiding deserialization overheads.
This leads us to our fourth observation.

\begin{table}[h]
\scalebox{0.8}{
\begin{tabular}{l|r|r|r}
\textbf{Pipeline} & \textbf{System-level} & \textbf{Application-level} & \textbf{Sample Size} \\ \hline
CV2-JPG           & $3.3\times$ & $15.2\times$ & 1.18\:MB \\ \hline
CV2-PNG           & $3.5\times$ & $14.5\times$ & 1.18\:MB \\ \hline
FLAC              & $4.2\times$ & $8.0\times$ & 0.41\:MB \\ \hline
MP3               & $1.6\times$ & $2.2\times$ & 0.08\:MB \\ \hline
NILM              & $1.1\times$ & $1.4\times$ & 0.01\:MB \\
\end{tabular}
}
\caption{{\color{diff}Throughput increase for different caching level compared to no caching of of each pipeline's last strategy.}}
\label{tab:throughput-caching-levels}
\end{table}
\vspace{-0.7cm}

\textbf{(4) Application-level caching is more efficient than system-level caching, but is still affected by the storage consumption per sample. }

}

{\color{diff2}

To understand how application-level caching affects the performance, we profiled all pipeline's respective last strategies, as well as our synthetic 15\:GB datasets again with application-level caching.
The results of the synthetic datasets in Fig.~\ref{fig:synthetic-dataset-cached-general} (\texttt{app-cache}) show that application-level caching is faster, but there is the same pattern of increasing processing time with a smaller sample size.
The online processing time with application-level caching consists solely of reading the samples from memory.
We can calculate the time spent on deserialization by subtracting the \texttt{app-cache} time from \texttt{sys-cache} time.
By dividing with the \texttt{sys-cache} time, we can get the percentage of the time spent on deserialization.
For the sample sizes 20.5\:MB to 5.1\:MB we spend 94-98\% time on deserialization ($\frac{4.8-0.1}{4.8}$), compared to 14-18\% for 0.08\:MB to 0.01\:MB ($\frac{167.3-138.3}{167.3}$).
Hence, the largest relative gains with application-level caching can be achieved with large sample sizes.

Our real-world pipelines have shown a similar throughput improvement compared to system-level caching when the dataset fits into memory (Tab.~\ref{tab:throughput-caching-levels}).
The decline in throughput improvement with both caching levels is directly correlated with a smaller sample size.
The last strategies of the CV and NLP pipelines failed to run with application-level caching as the dataset did not fit into the cache (cf. Sec.~\ref{ssec:caching} \textbf{(1)}).

}

{\color{diff}

\subsection{Compression}
\label{ssec:compression}

Storage consumption has shown to be an important factor for throughput.
Compression adds a new possibility to decrease storage consumption at the cost of an offline compression step and an online decompression step.
For compression to provide a benefit, the gains of decreased data size must outweigh the computational overheads.
A common metric to evaluate compression on storage consumption is the \textit{space saving} percentage.
For example, if the size did not change after compression, the space saving is 0\%.
When it changes from originally 5\:GB to 1\:GB, the space saving is 80\%.
We omitted the \texttt{unprocessed} strategy for all pipelines because accessing single files is bound by the random access performance of the storage (cf. Sec.~\ref{ssec:storage-versus-throughput} \textbf{(1)}) and compression does not help with this issue.
\begin{figure}
    \begin{subfigure}[c]{0.26\textwidth}
        \includegraphics[width=\textwidth]{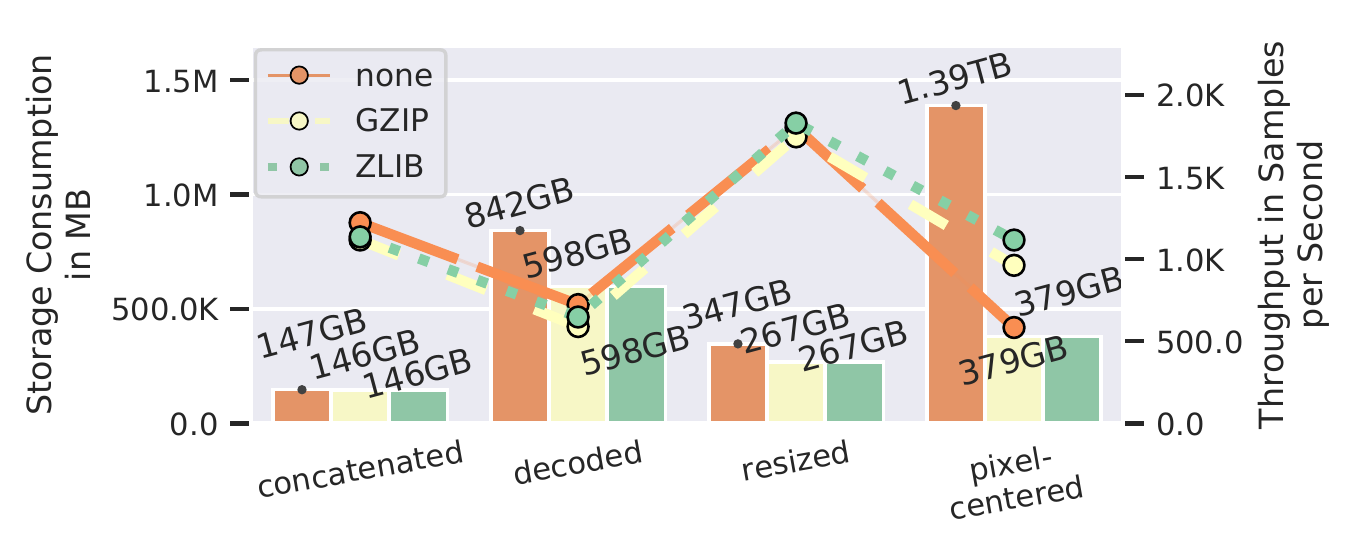}
        \vspace{-18pt}
        \caption{CV\footnotemark[1]}
        \label{fig:compressed-storage-vs-throughput-cv}
    \end{subfigure}
    \begin{subfigure}[c]{0.21\textwidth}
        \includegraphics[width=\textwidth]{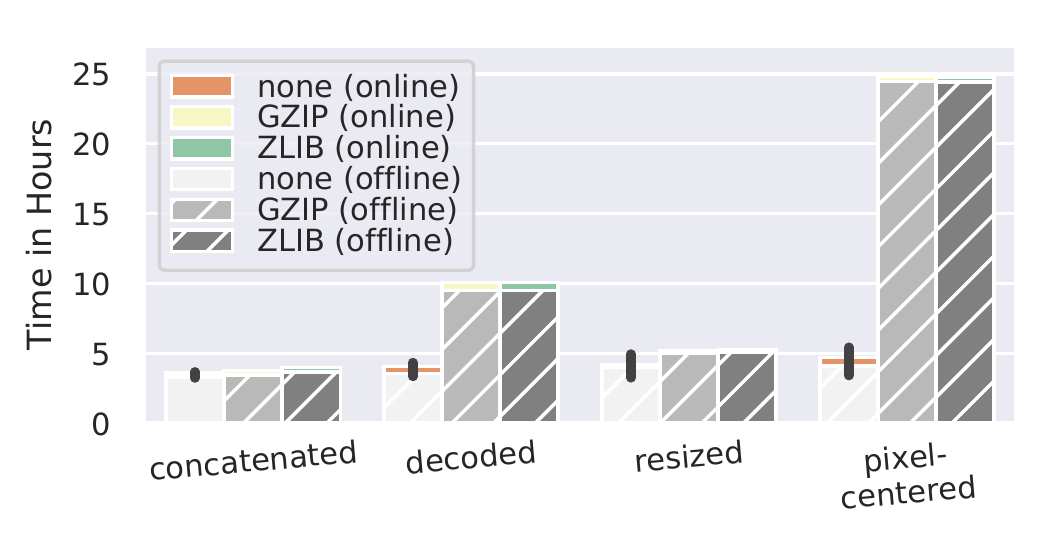}
        \vspace{-18pt}
        \caption{CV\footnotemark[1]}
        \label{fig:compressed-processing-time-cv}
    \end{subfigure}
    \begin{subfigure}[c]{0.26\textwidth}
        \includegraphics[width=\textwidth]{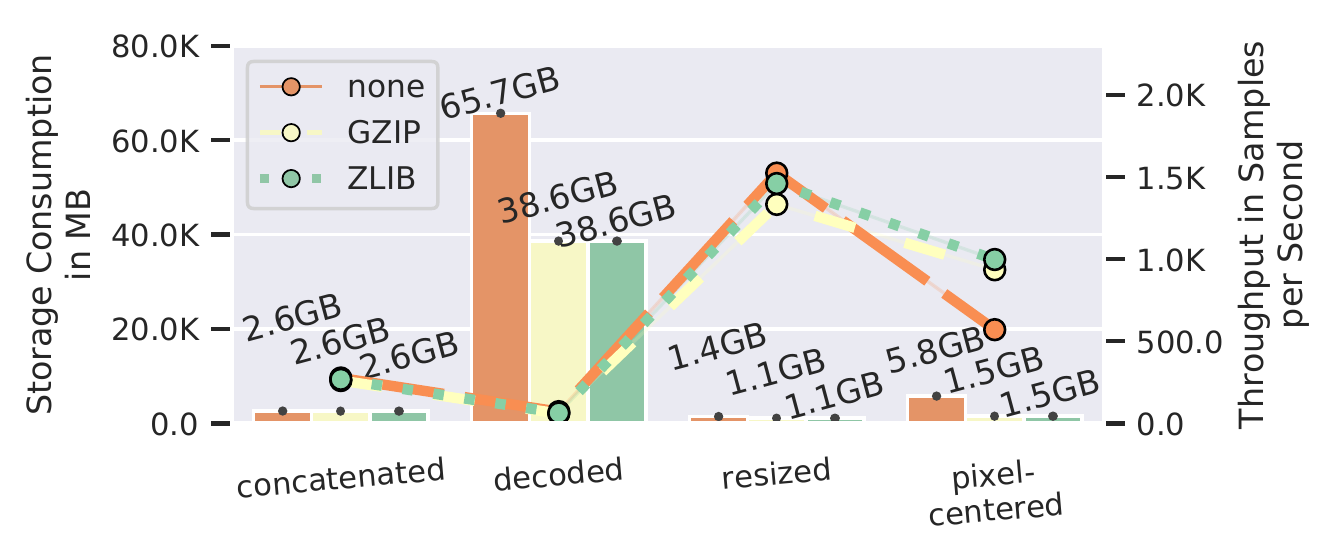}
        \vspace{-18pt}
        \caption{CV2-JPG}
        \label{fig:compressed-storage-vs-throughput-cv2-jpg}
    \end{subfigure}
    \begin{subfigure}[c]{0.21\textwidth}
        \includegraphics[width=\textwidth]{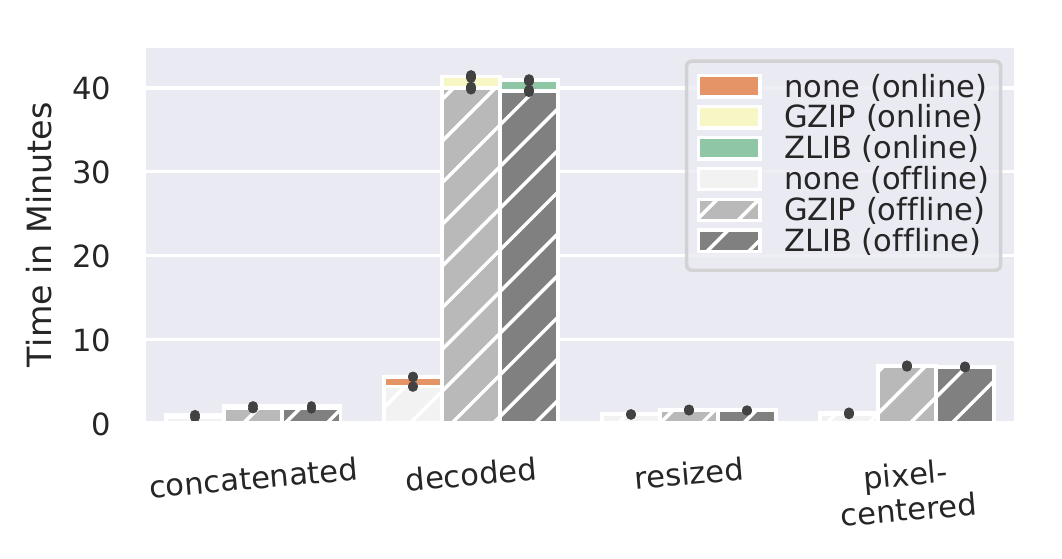}
        \vspace{-18pt}
        \caption{CV2-JPG}
        \label{fig:compressed-processing-time-cv2-jpg}
    \end{subfigure}
    \begin{subfigure}[c]{0.26\textwidth}
        \includegraphics[width=\textwidth]{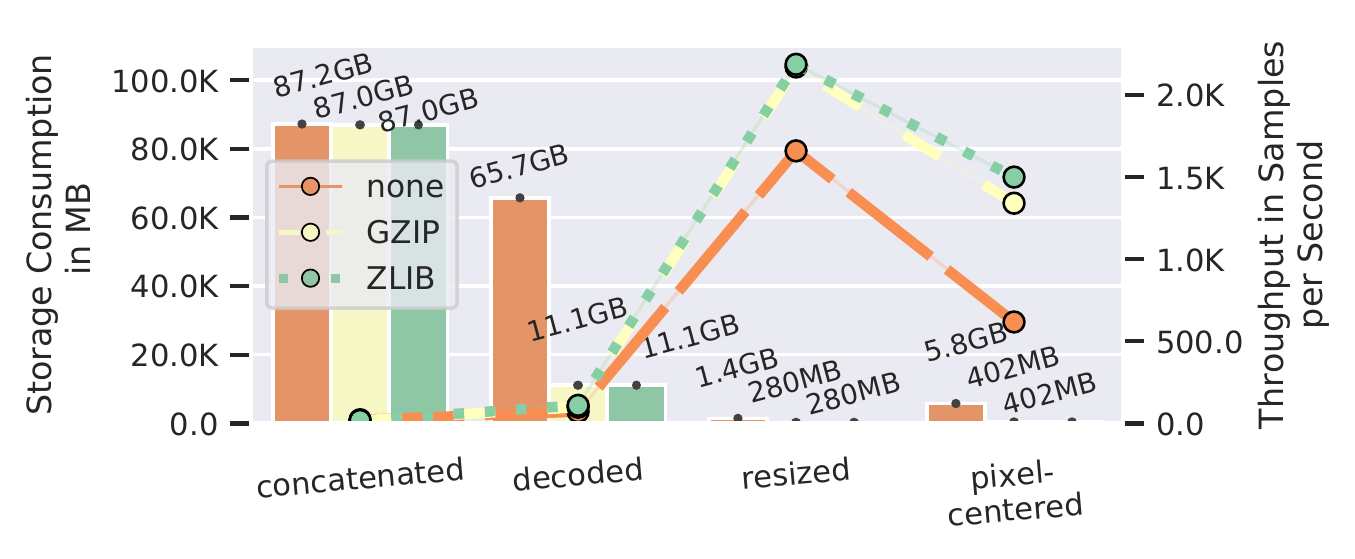}
        \vspace{-18pt}
        \caption{CV2-PNG}
        \label{fig:compressed-storage-vs-throughput-cv2-png}
    \end{subfigure}
    \begin{subfigure}[c]{0.21\textwidth}
        \includegraphics[width=\textwidth]{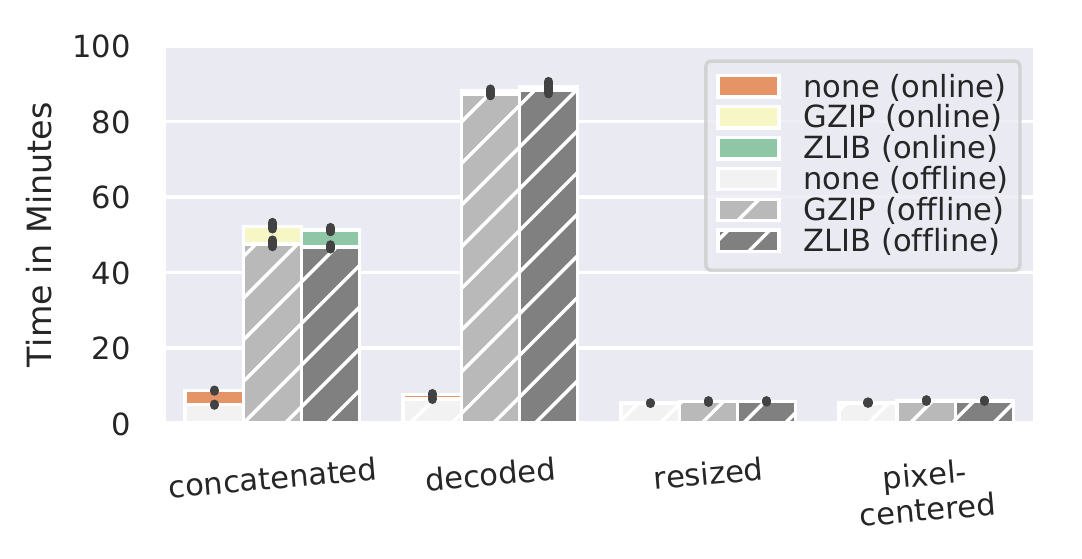}
        \vspace{-18pt}
        \caption{CV2-PNG}
        \label{fig:compressed-processing-time-cv2-png}
    \end{subfigure}
    \begin{subfigure}[c]{0.26\textwidth}
        \includegraphics[width=\textwidth]{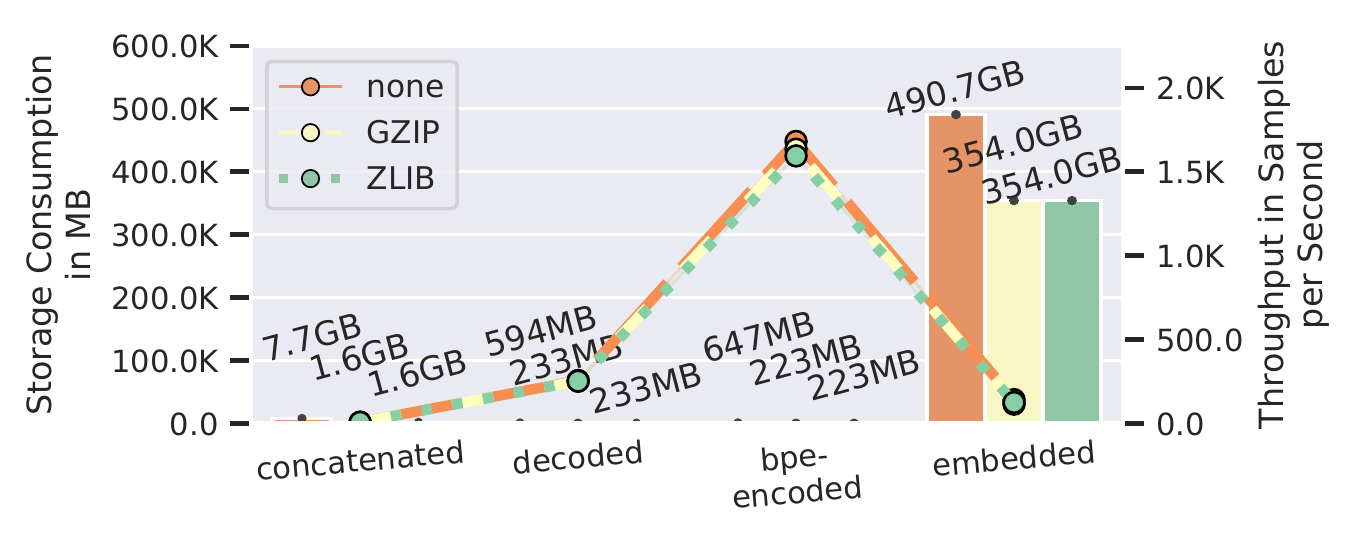}
        \vspace{-18pt}
        \caption{NLP}
        \label{fig:compressed-storage-vs-throughput-nlp}
    \end{subfigure}
    \begin{subfigure}[c]{0.21\textwidth}
        \includegraphics[width=\textwidth]{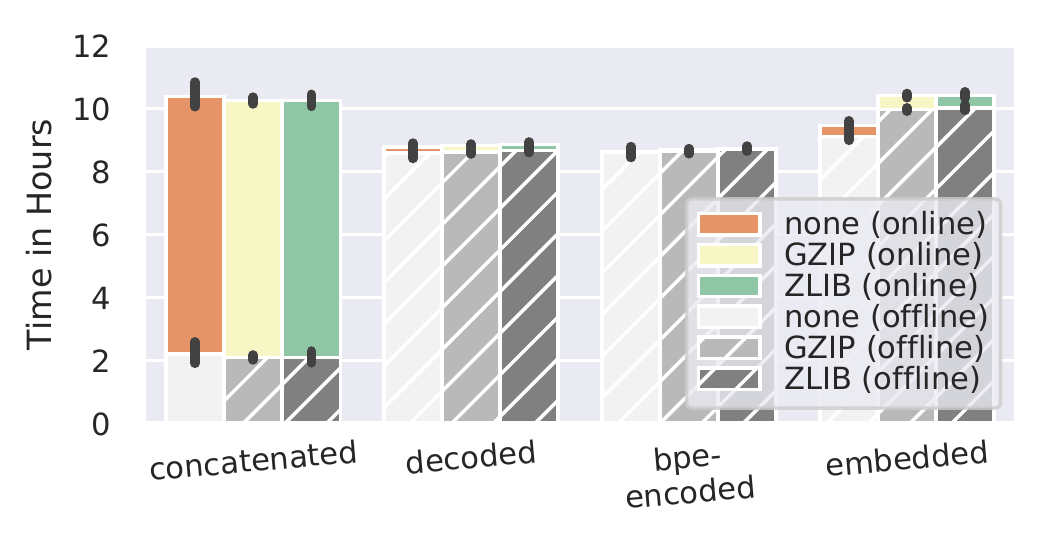}
        \vspace{-18pt}
        \caption{NLP}
        \label{fig:compressed-processing-time-nlp}
    \end{subfigure}
    \begin{subfigure}[c]{0.26\textwidth}
        \includegraphics[width=\textwidth]{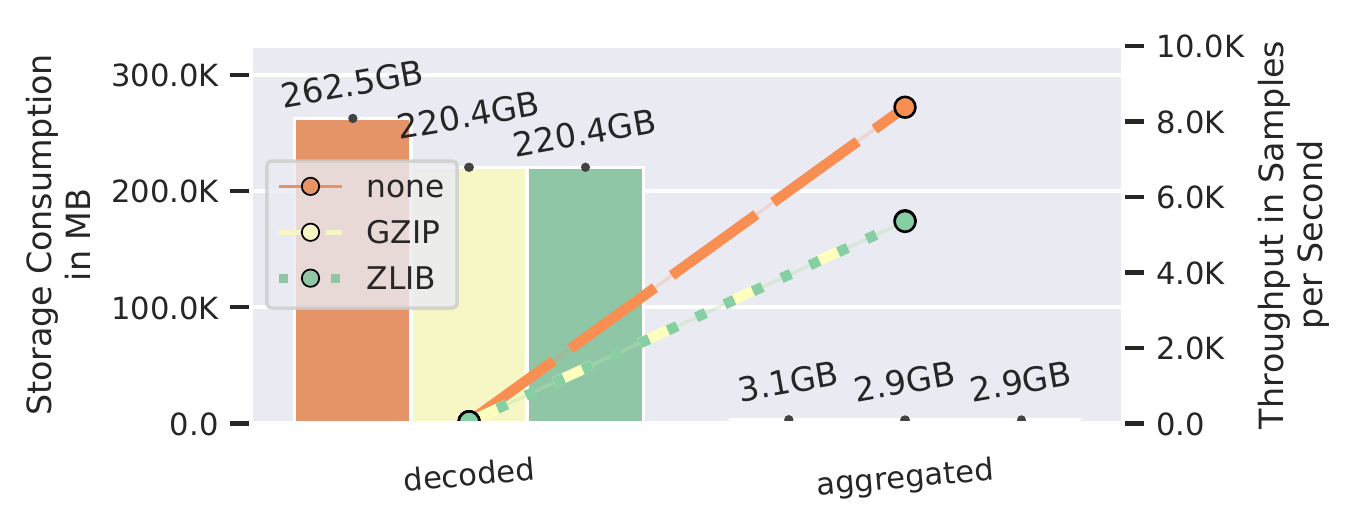}
        \vspace{-18pt}
        \caption{NILM}
        \label{fig:compressed-storage-vs-throughput-nilm}
    \end{subfigure}
    \begin{subfigure}[c]{0.21\textwidth}
        \includegraphics[width=\textwidth]{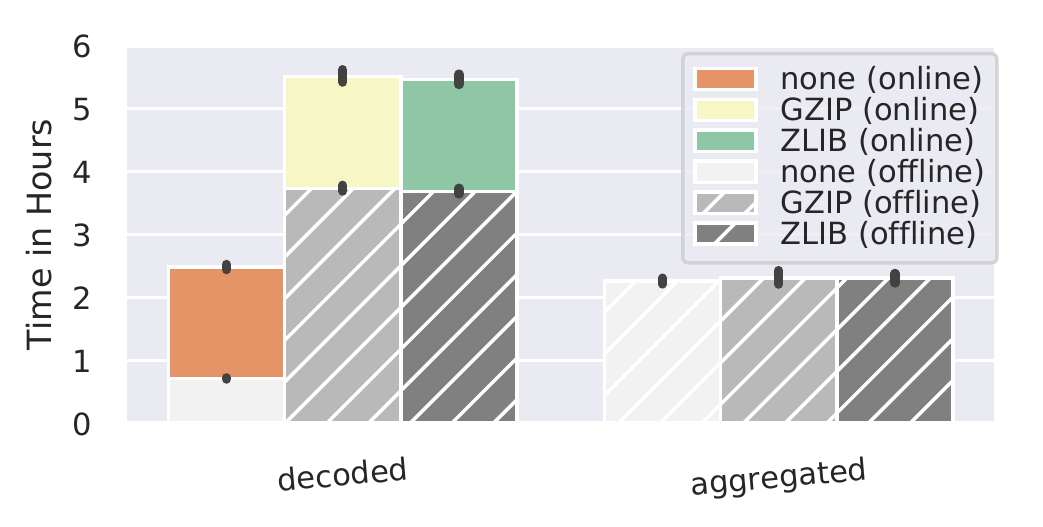}
        \vspace{-18pt}
        \caption{NILM}
        \label{fig:compressed-processing-time-nilm}
    \end{subfigure}
    \begin{subfigure}[c]{0.26\textwidth}
        \includegraphics[width=\textwidth]{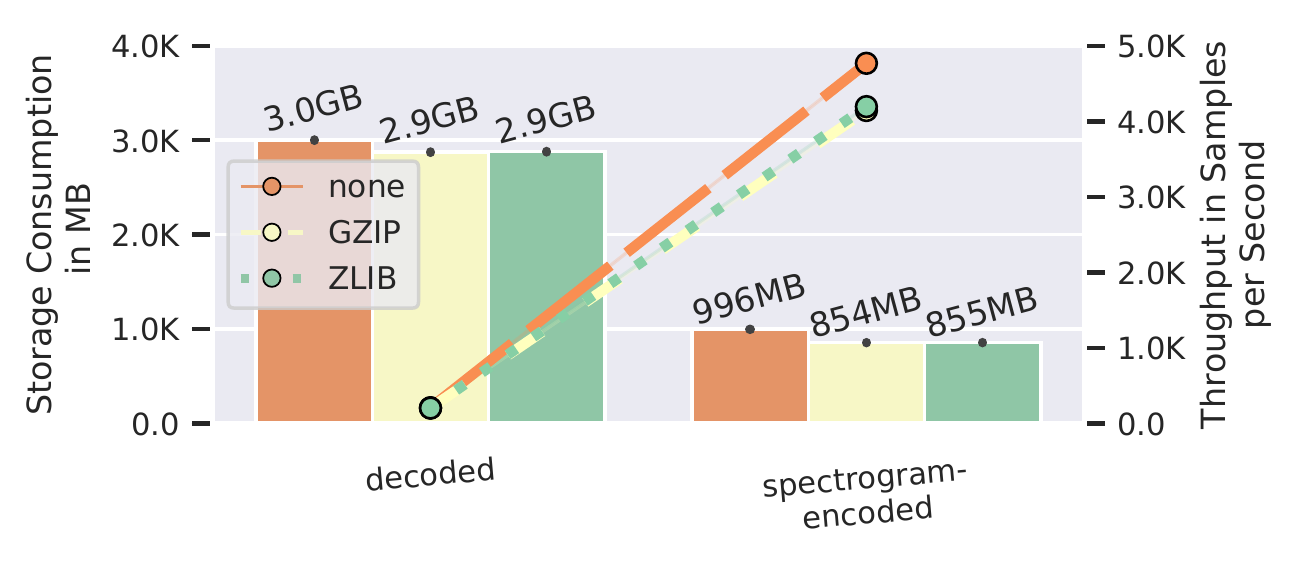}
        \vspace{-18pt}
        \caption{MP3}
        \label{fig:compressed-storage-vs-throughput-mp3}
    \end{subfigure}
        \begin{subfigure}[c]{0.21\textwidth}
        \includegraphics[width=\textwidth]{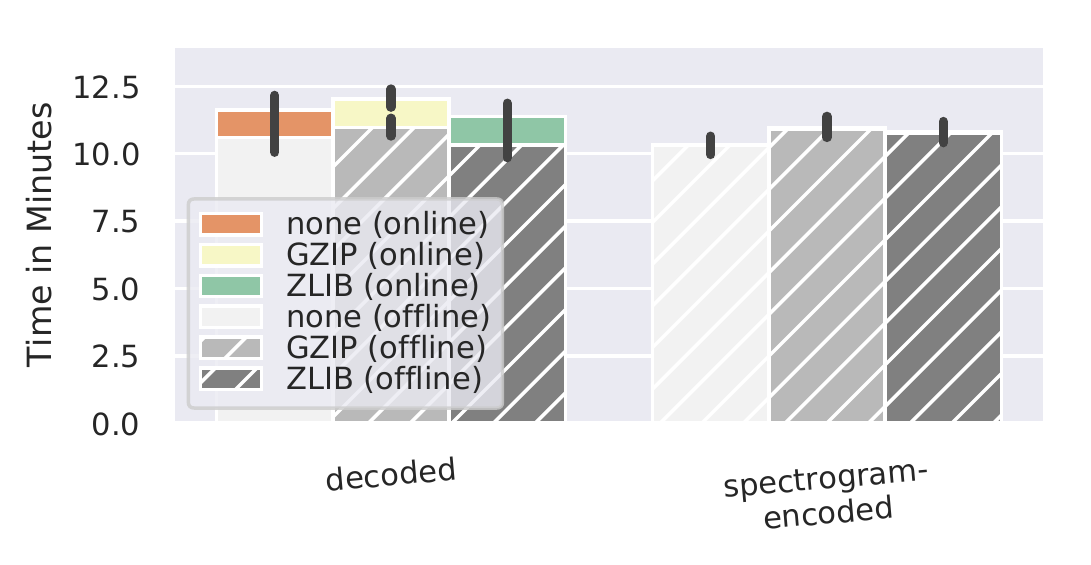}
        \vspace{-18pt}
        \caption{MP3}
        \label{fig:compressed-processing-time-mp3}
    \end{subfigure}    
    \begin{subfigure}[c]{0.26\textwidth}
        \includegraphics[width=\textwidth]{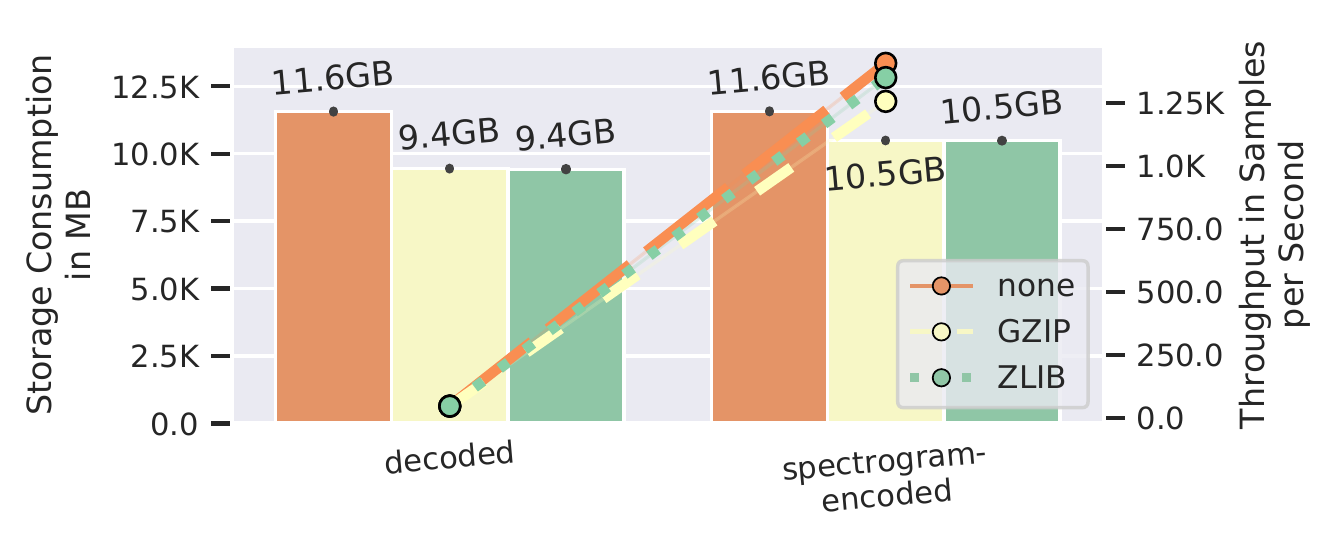}
        \vspace{-18pt}
        \caption{FLAC}
        \label{fig:compressed-storage-vs-throughput-flac}
    \end{subfigure}
    \begin{subfigure}[c]{0.21\textwidth}
        \includegraphics[width=\textwidth]{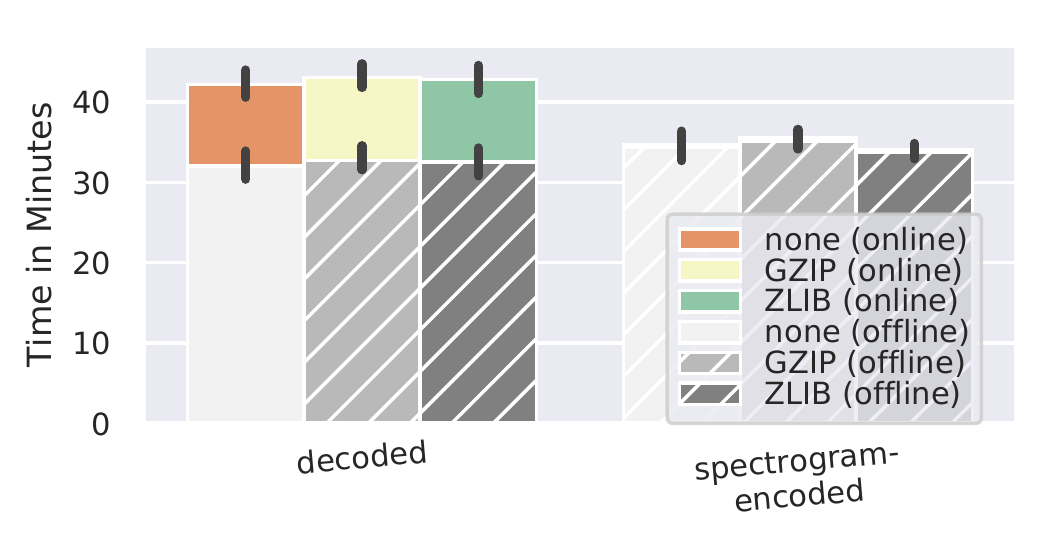}
        \vspace{-18pt}
        \caption{FLAC}
        \label{fig:compressed-processing-time-flac}
    \end{subfigure}
    \vspace{-0.3cm}
    \caption{{\color{diff}Left column: Storage consumption compared to $T_4$ throughput (dotted lines) with compression. Right column: Offline (grey hatched bars) and online processing time (colored) with compression.}}
    \label{fig:compression-results}
\end{figure}
\footnotetext[1]{Unfortunately, after one full run with each compression library (GZIP, ZLIB) for each strategy of the CV pipeline, our CEPH storage system was reconfigured which led to non-comparable results of the respective repeat runs. Hence, for these specific experiments, we only report results of one run (instead of the average of five runs, as in all other experiments).}
The results of our compression experiments are shown in Fig.~\ref{fig:compression-results}.
We make the following observations:

\textbf{(1) High space savings do not guarantee improved throughput.}

Space saving affects the throughput positively for some, but not all strategies.
All CV-based pipelines had an increase in throughput with compression between $1.6\times$ and $2.4\times$ at \texttt{pixel-centered} where space saving is between 73\% and 93\% (Fig.~\ref{fig:compressed-storage-vs-throughput-cv},~\ref{fig:compressed-storage-vs-throughput-cv2-jpg},~\ref{fig:compressed-storage-vs-throughput-cv2-png}).
In this case, the faster read time in total was beneficial compared to the cost of the additional decompression step.

In contrast, the strategies of the NLP pipeline have a space saving between 28\% and 80\%, but none of them had a throughput increase (Fig.~\ref{fig:compressed-storage-vs-throughput-nlp}).
The reason for that is that every strategy was bound by a computationally expensive CPU step except the last strategy \texttt{embedded}.
At \texttt{embedded}, the dataset is only read from disk and deserialized, but the space saving of 28\% was not enough to benefit the total throughput.

The same effect becomes visible when comparing the strategies \texttt{decoded} and \texttt{resized} between the CV2-PNG and CV2-JPG pipelines (Fig.~\ref{fig:compressed-storage-vs-throughput-cv2-jpg},~\ref{fig:compressed-storage-vs-throughput-cv2-png}).
The only difference between the pipelines is the encoding of the images, with JPG being a lossy storage format, while PNG is lossless.
The CV2-PNG pipeline has a better space saving with compression with the \texttt{decoded} strategy (83\%) compared to CV2-JPG (41\%).
This results in a throughput increase by $1.5\times$ for CV2-PNG compared to the 89\% throughput deterioration at CV2-JPG.
The \texttt{resized} strategy with the PNG images has a space saving of 81\% and improves the throughput by $1.3\times$, while JPG only saves 24\% of space and reduces the throughput to 96\%.
The compression artifacts introduced by the lossy JPG encoding affect the space saving of both GZIP and ZLIB negatively.

All the other pipelines, NILM, MP3 and FLAC, slow down with compression and have a varying space saving between 0.3-41.2\% (Fig.~\ref{fig:compressed-storage-vs-throughput-nilm},~\ref{fig:compressed-storage-vs-throughput-mp3},~\ref{fig:compressed-storage-vs-throughput-flac}).
When comparing the different compression types, ZLIB was slightly faster and had a comparable space saving to GZIP, except for NLP's \texttt{bpe-encoded}, where it was slightly slower compared to GZIP.

\textbf{(2) Offline compression and write time can be volatile. }

When compressing a dataset, the processing time is increased by the compression algorithm, and decreased by the lower write time due to lower storage consumption.
The balance between these steps is not predictable from our observations, as the CV2-PNG pipeline shows (Fig.~\ref{fig:compressed-processing-time-cv2-png}). 
With the \texttt{concatenated} strategy and a space saving of only 0.3\%, it takes $9.6\times$ longer to save the dataset to storage.
The strategy \texttt{decoded} has a space saving of 83\% and takes $13.5\times$ longer for offline processing.
The next strategies, \texttt{resized} and \texttt{pixel-centered}, have a space saving of 80\%-93\% and only take 1.08-1.1$\times$ longer.
Compared to a slightly worse space saving of 74\% with the \texttt{pixel-centered} strategy at the CV2-JPG pipeline (Fig.~\ref{fig:compressed-processing-time-cv2-jpg}), the offline processing time is increased by $6.1\times$. 

Generally, we see examples of a high space saving and no effective increase in offline processing time in NLP (Fig.~\ref{fig:compressed-processing-time-nlp}), a low space saving with a higher offline processing time in NILM (Fig.~\ref{fig:compressed-processing-time-nilm}) and CV2-PNG \texttt{concatenated} (Fig.~\ref{fig:compressed-processing-time-cv2-png}), and low space saving with no effective increase in offline processing time in MP3 (Fig.~\ref{fig:compressed-processing-time-mp3}), FLAC (Fig.~\ref{fig:compressed-processing-time-flac}), and CV \texttt{concatenated} and \texttt{resized} (Fig.~\ref{fig:compressed-processing-time-cv}). Space saving does not seem to be a good predictor at how the compression will affect the offline processing time.

}

{\color{diff}

\subsection{Parallelization Capabilities}
\label{ssec:parallelization-capabilities}

\vspace{-0.45cm}
\begin{figure}[h]
    \centering
    \includegraphics[width=0.47\textwidth]{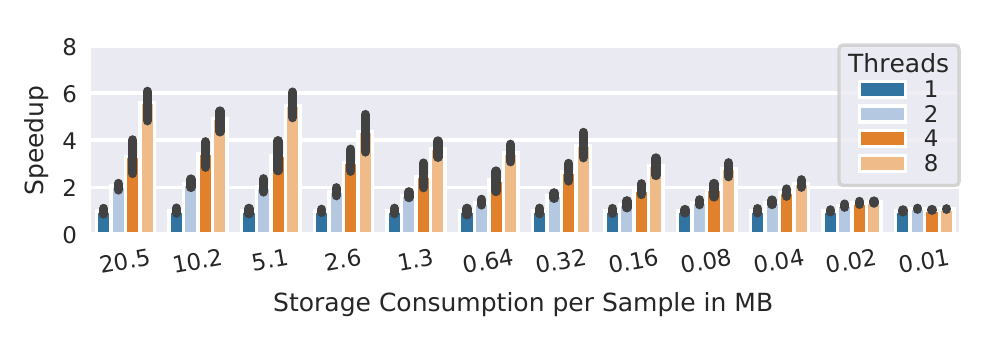}
    \vspace{-15pt}
    \caption{{\color{diff}Reading a synthetic 15\:GB dataset with different sample sizes to compare multi-threaded scalability.}}
    \label{fig:synthetic-dataset-speedup}
\end{figure}
\vspace{-0.2cm}

\begin{figure}
    \begin{subfigure}[c]{0.22\textwidth}
        \includegraphics[width=\textwidth]{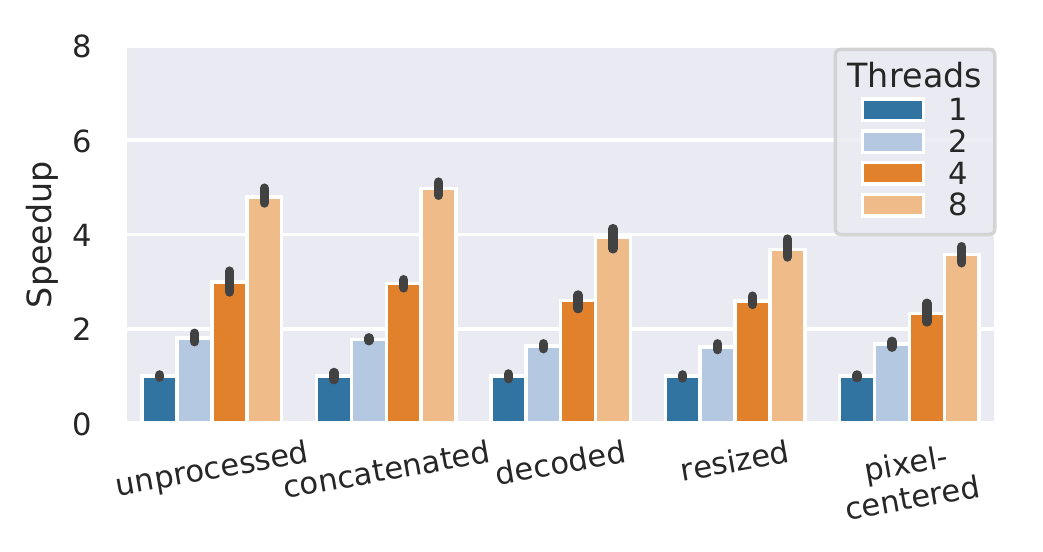}
        \vspace{-18pt}
        \caption{CV \texttt{no-cache}}
        \label{fig:speedup-cv}
    \end{subfigure}
    \begin{subfigure}[c]{0.22\textwidth}
        \includegraphics[width=\textwidth]{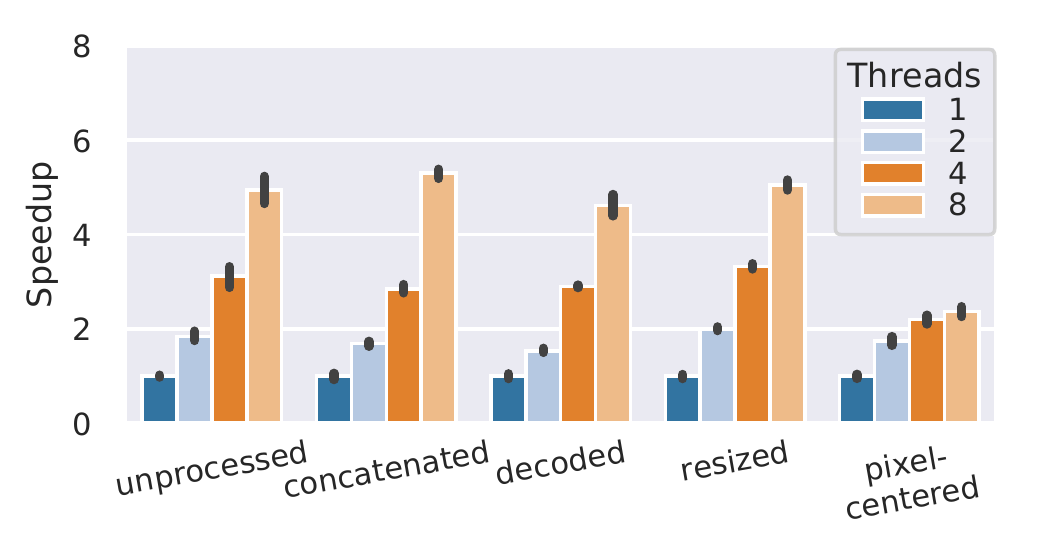}
        \vspace{-18pt}
        \caption{CV \texttt{sys-cache}}
        \label{fig:speedup-epochs-cv}
    \end{subfigure}
    \begin{subfigure}[c]{0.22\textwidth}
        \includegraphics[width=\textwidth]{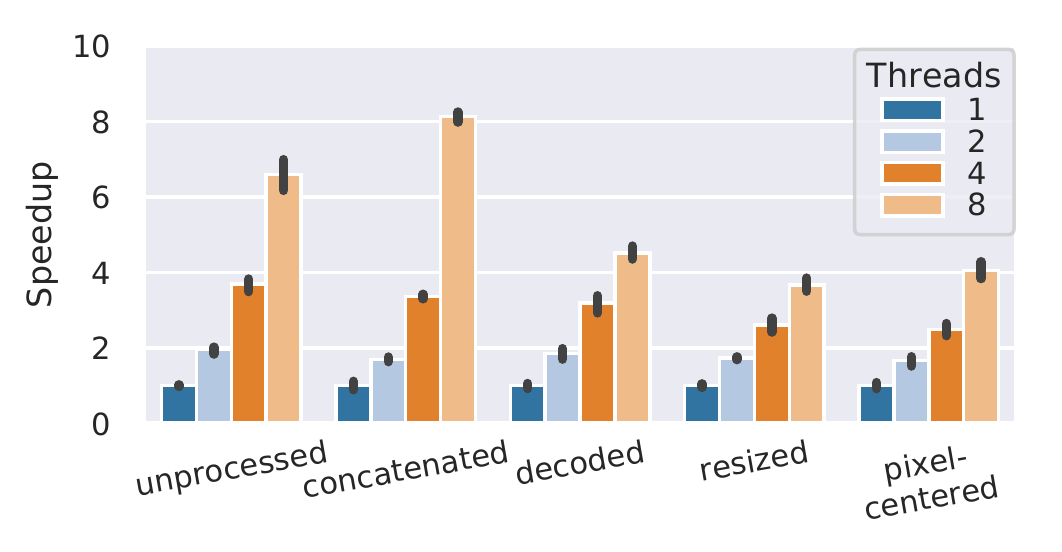}
        \vspace{-18pt}
        \caption{CV2-JPG \texttt{no-cache}}
        \label{fig:speedup-cv2-jpg}
    \end{subfigure}
    \begin{subfigure}[c]{0.22\textwidth}
        \includegraphics[width=\textwidth]{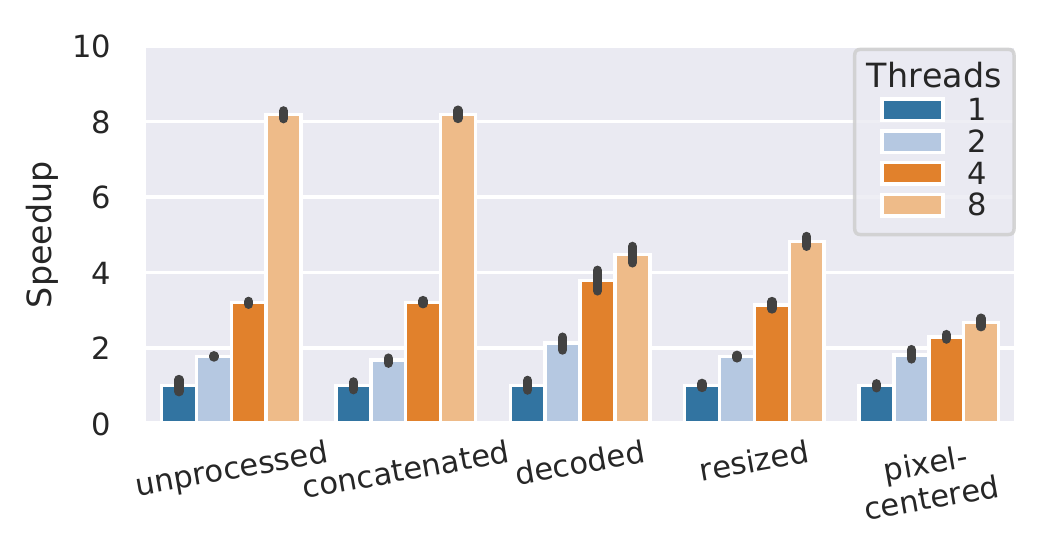}
        \vspace{-18pt}
        \caption{CV2-JPG \texttt{sys-cache}}
        \label{fig:fig:speedup-epochs-cv2-jpg}
    \end{subfigure}
    \begin{subfigure}[c]{0.22\textwidth}
        \includegraphics[width=\textwidth]{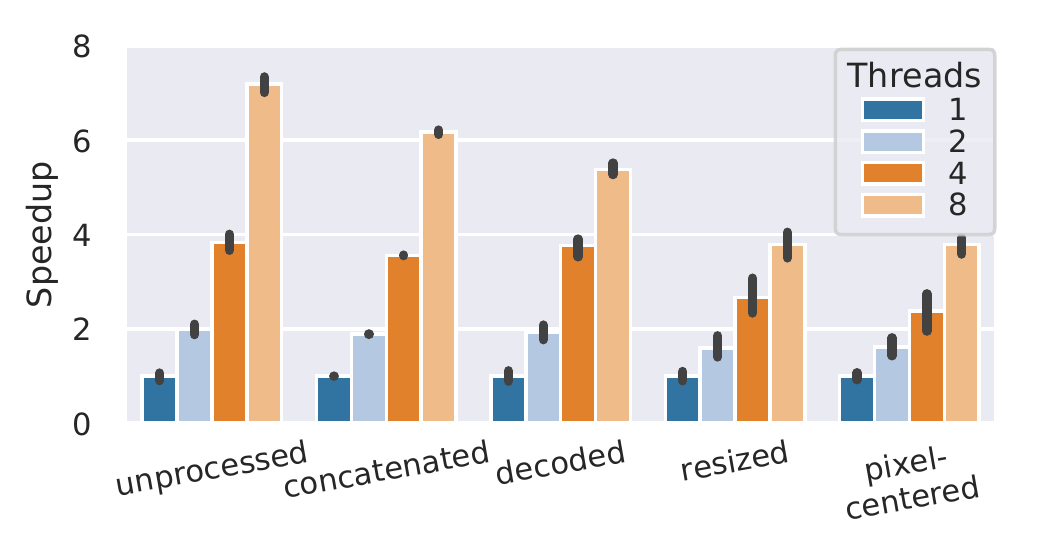}
        \vspace{-18pt}
        \caption{CV2-PNG \texttt{no-cache}}
        \label{fig:speedup-cv2-png}
    \end{subfigure}
    \begin{subfigure}[c]{0.22\textwidth}
        \includegraphics[width=\textwidth]{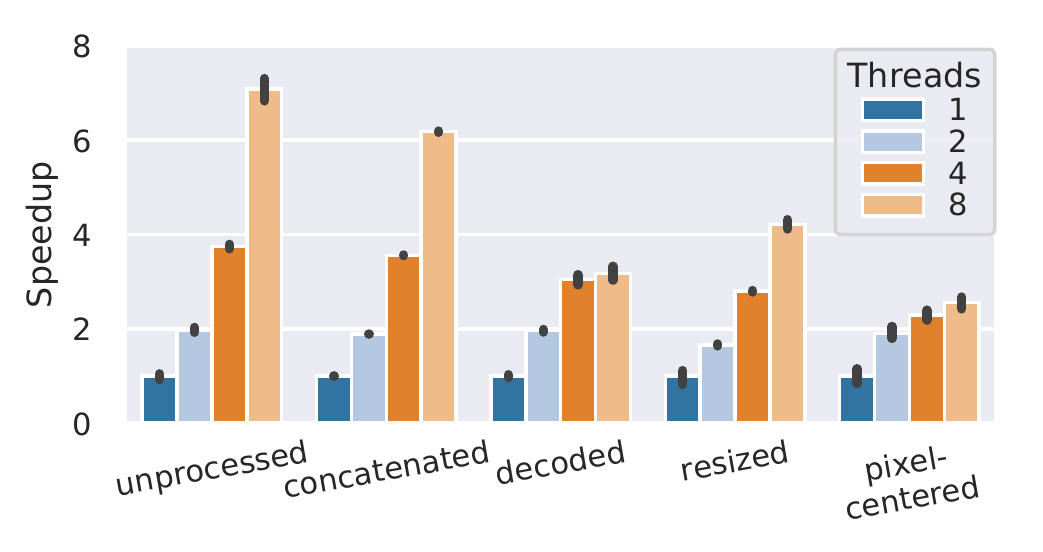}
        \vspace{-18pt}
        \caption{CV2-PNG \texttt{sys-cache}}
        \label{fig:speedup-epochs-cv2-png}
    \end{subfigure}
    \begin{subfigure}[c]{0.22\textwidth}
        \includegraphics[width=\textwidth]{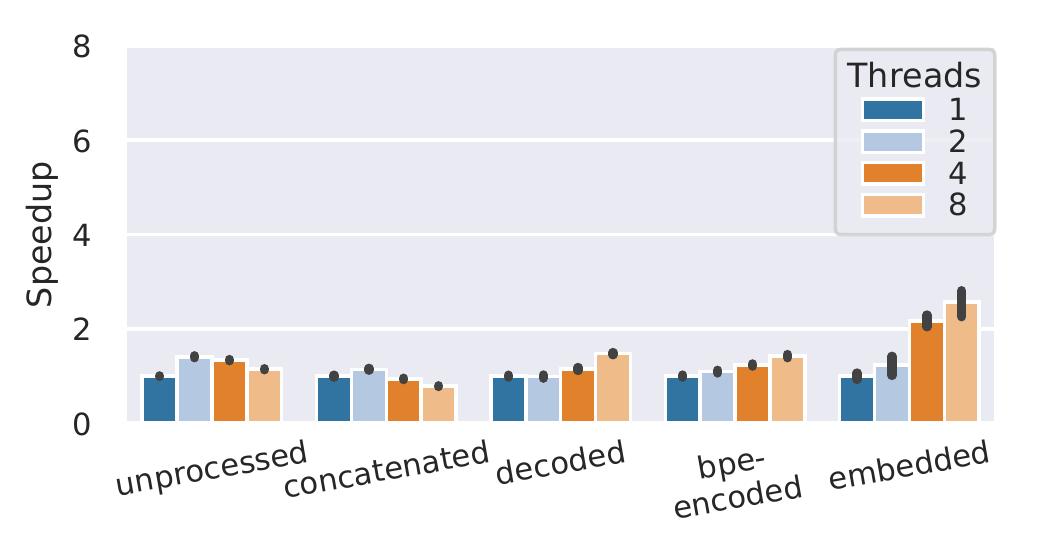}
        \vspace{-18pt}
        \caption{NLP \texttt{no-cache}}
        \label{fig:speedup-nlp}
    \end{subfigure}
    \begin{subfigure}[c]{0.22\textwidth}
        \includegraphics[width=\textwidth]{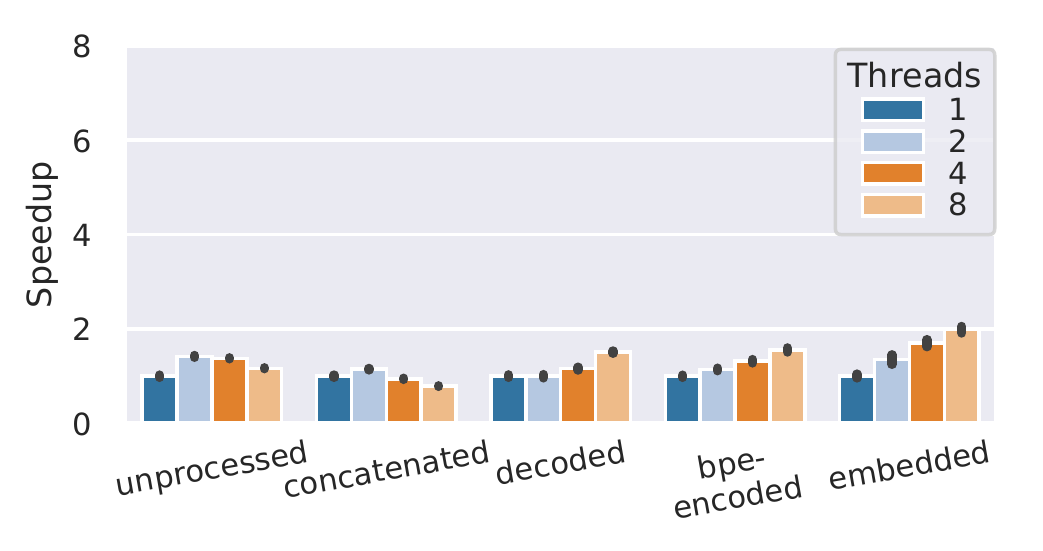}
        \vspace{-18pt}
        \caption{NLP \texttt{sys-cache}}
        \label{fig:speedup-epochs-nlp}
    \end{subfigure}
    \begin{subfigure}[c]{0.22\textwidth}
        \includegraphics[width=\textwidth]{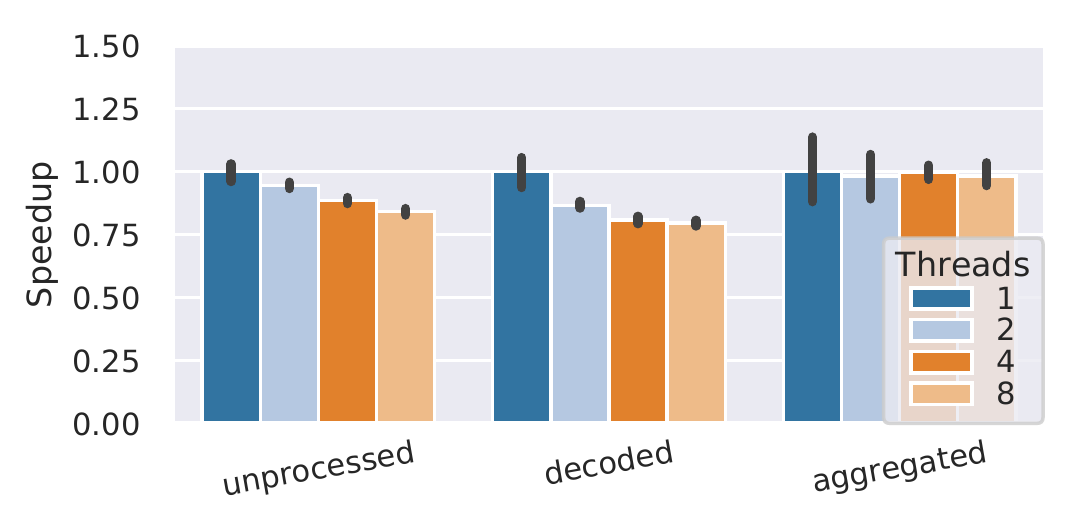}
        \vspace{-18pt}
        \caption{NILM \texttt{no-cache}}
        \label{fig:speedup-nilm}
    \end{subfigure}
    \begin{subfigure}[c]{0.22\textwidth}
        \includegraphics[width=\textwidth]{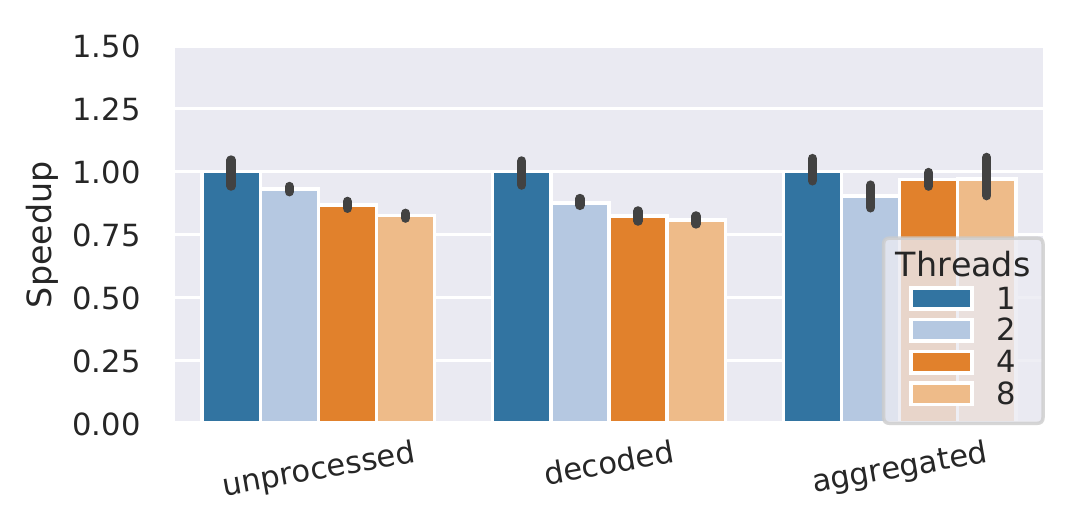}
        \vspace{-18pt}
        \caption{NILM \texttt{sys-cache}}
        \label{fig:speedup-epochs-nilm}
    \end{subfigure}
    \begin{subfigure}[c]{0.22\textwidth}
        \includegraphics[width=\textwidth]{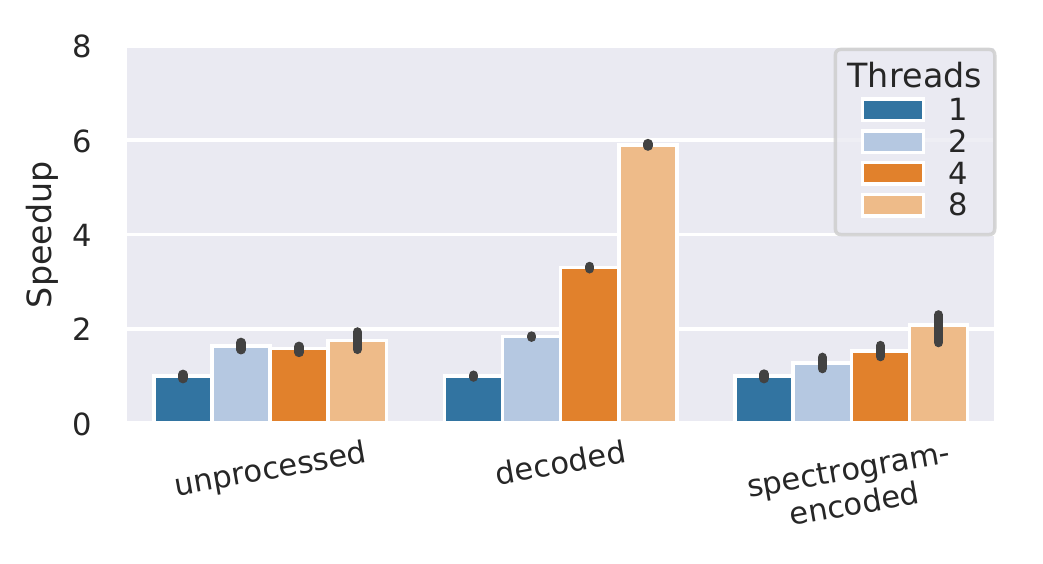}
        \vspace{-18pt}
        \caption{MP3 \texttt{no-cache}}
        \label{fig:speedup-mp3}
    \end{subfigure}
        \begin{subfigure}[c]{0.22\textwidth}
        \includegraphics[width=\textwidth]{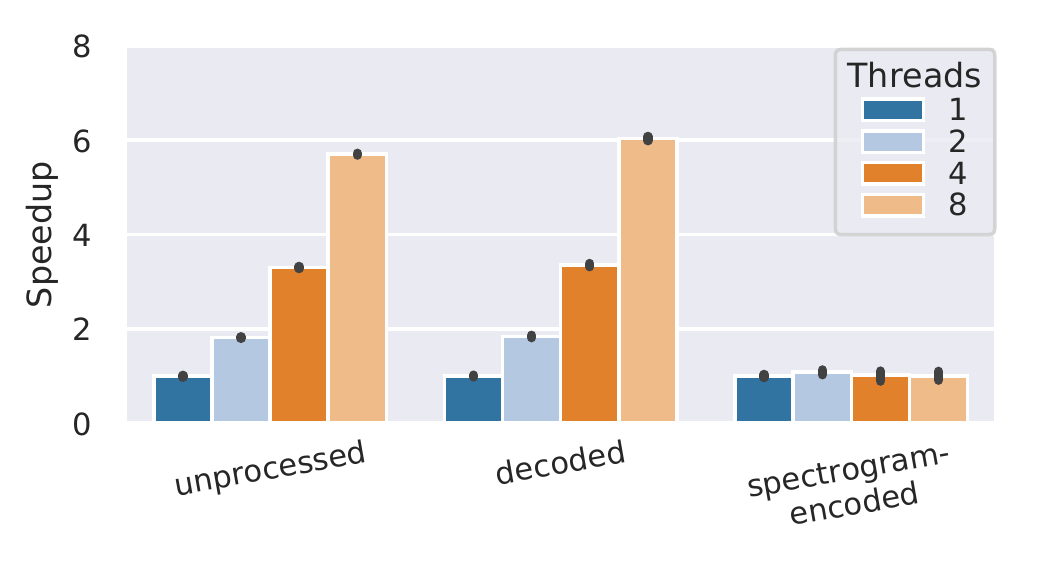}
        \vspace{-18pt}
        \caption{MP3 \texttt{sys-cache}}
        \label{fig:speedup-epochs-mp3}
    \end{subfigure}
    \begin{subfigure}[c]{0.22\textwidth}
        \includegraphics[width=\textwidth]{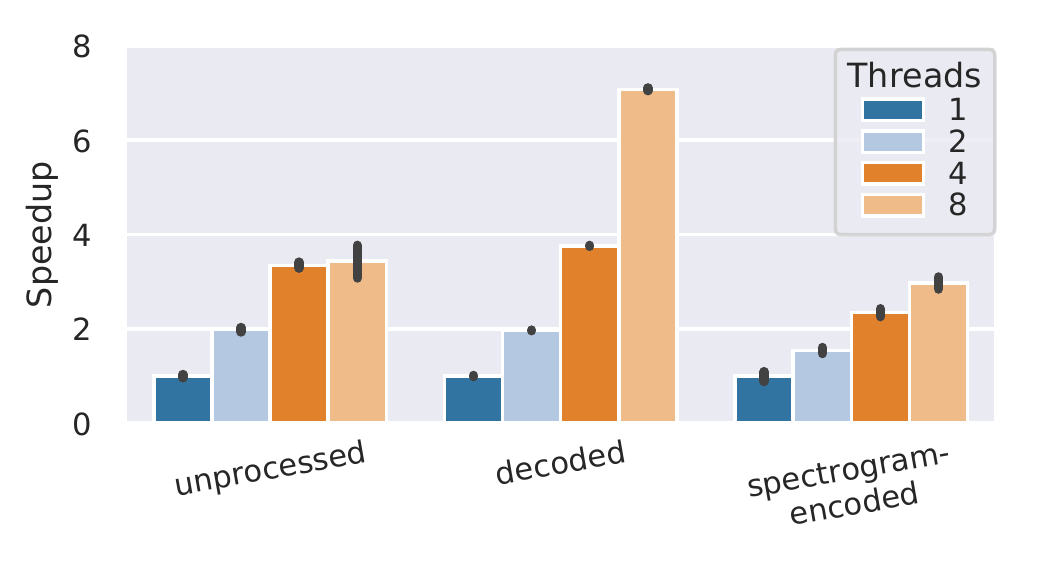}
        \vspace{-18pt}
        \caption{FLAC \texttt{no-cache}}
        \label{fig:speedup-flac}
    \end{subfigure}
    \begin{subfigure}[c]{0.22\textwidth}
        \includegraphics[width=\textwidth]{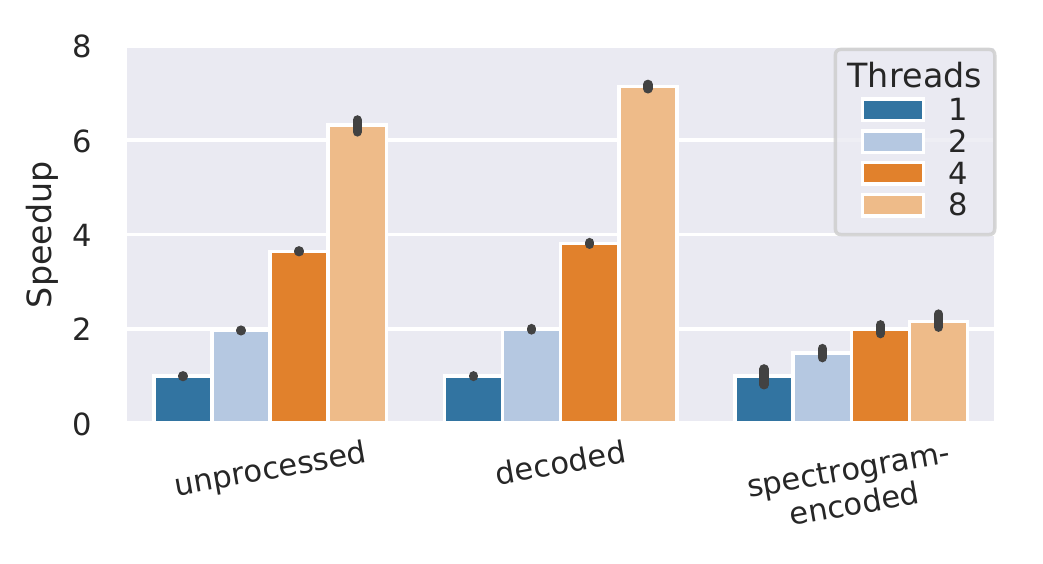}
        \vspace{-18pt}
        \caption{FLAC \texttt{sys-cache}}
        \label{fig:speedup-epochs-flac}
    \end{subfigure}
    \vspace{-0.3cm}
    \caption{{\color{diff}Speedup at 8000 samples. Left column: No caching. Right column: System-level caching.}}
    \label{fig:speedup}
\end{figure}

We analyzed the parallelization capabilities of each pipeline under multi-threading, as this is one of the best practice recommendations to speed up data pipelines and remove I/O bottlenecks~\cite{tfdatabestpractice2020}.
We compared the speedup of each strategy by running it with 1, 2, 4, and 8 threads over two epochs with system-level caching enabled (Fig.~\ref{fig:speedup}).
The profiling was done with a fraction of the dataset (up to 8000 samples) so that the second epoch could be fully cached for each pipeline to compare the speedup with system-level caching.
We make the following observations:

\textbf{(1) A small storage consumption per sample hinders multi-threaded performance.}
We know from the previous Sections~\ref{ssec:storage-versus-throughput} and \ref{ssec:caching} that a small storage consumption per sample affects the online preprocessing time negatively.
However, how does it affect multi-threaded execution?
To analyze this, we reused the same synthetic 15\:GB \texttt{float32} dataset with different sample sizes to compare their multi-threaded read and deserialization time.
The results in Fig.~\ref{fig:synthetic-dataset-speedup} show a similar trend as before, with a speedup of close to 1$\times$ for the 0.01\:MB sample sizes.
This means that processing small samples with a single thread takes equally long as with eight threads.
}
{\color{diff2}
We traced the issue down to an increased amount of context switches with smaller sample sizes (100,000 per second at 0.01\:MB compared to 5,000 per second for 20.5\:MB).
Additionaly, as we extracted from the trace log, every thread only processes a single sample at a time and is finished faster with smaller sample sizes before being scheduled again.
Scheduling a thread to process a new sample induces so much overhead that multi-threading can not be effective at small sample sizes.
}

{\color{diff}

A good example from our real-world pipelines is NILM (Fig.~\ref{fig:speedup-nilm}) at the \texttt{aggregated} strategy, which has no effective speedup due to a sample size of 0.01\:MB.
Even when reading the dataset from memory (Fig.~\ref{fig:speedup-epochs-nilm}), there is virtually no change in speedup.
Every last strategy from each pipeline has as slightly worse speedup when reading from memory (\texttt{sys-cache}) than when reading from storage (\texttt{no-cache}) (Fig.~\ref{fig:speedup}).
The reason for this is that memory provides a higher bandwidth, so reading data is fast, even with a single thread.
Therefore, the effect of context switches is highlighted even more compared to the case where the slower network read speeds affects the total processing time additionally.

\newpage
\textbf{(2) Inefficient preprocessing can reduce multi-threading scalability. }

While most of the scaling issues like \texttt{bpe-encoding} in NLP can be explained with a small sample size (0.003\:MB), we also observed slowdowns (speedup $< 1.0$), which have a different root cause.
This happens with the first two strategies of NILM (Fig.~\ref{fig:speedup-nilm}, 0.15\:MB and 0.98\:MB) and NLP (Fig.~\ref{fig:speedup-nlp}, 0.04MB), which are not alleviated by reading from memory (Fig.~\ref{fig:speedup-epochs-nilm}, ~\ref{fig:speedup-epochs-nlp}, respectively).
This points to a processing issue.
One thing that both \texttt{unprocessed}, \texttt{concatenated} (NLP) and \texttt{decoded} (NILM) have in common, is that they are using external Python libraries like NumPy and newspaper wrapped in a \texttt{tf.py\_function}, while all the other preprocessing steps are provided by the TensorFlow library.

}

{\color{diff2}
\vspace{-0.3cm}
\begin{figure}[h]
    \begin{subfigure}[c]{0.23\textwidth}
        \includegraphics[width=\textwidth]{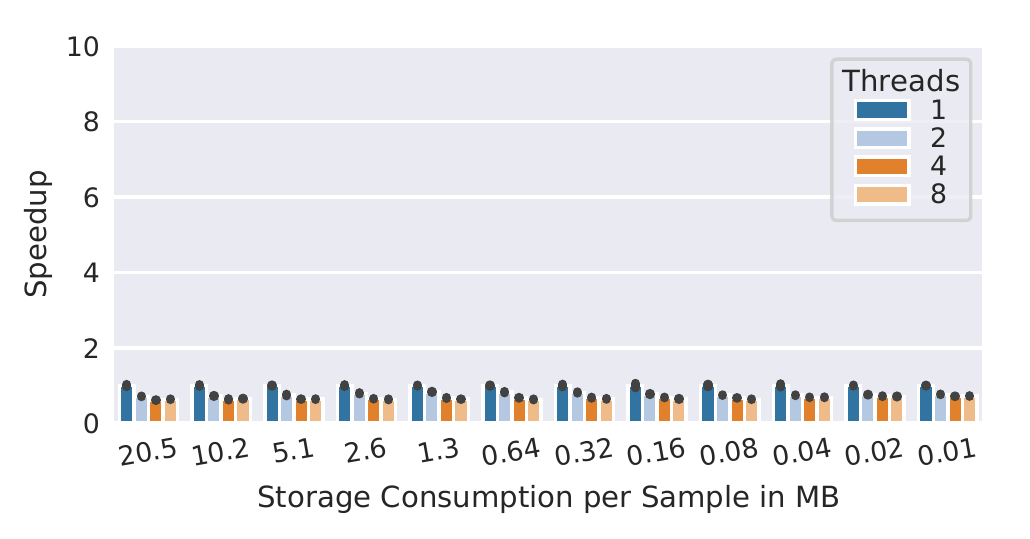}
        \vspace{-18pt}
        \caption{NumPy RMS}
        \label{fig:synthetic-dataset-processing-np}
    \end{subfigure}
    \begin{subfigure}[c]{0.23\textwidth}
        \includegraphics[width=\textwidth]{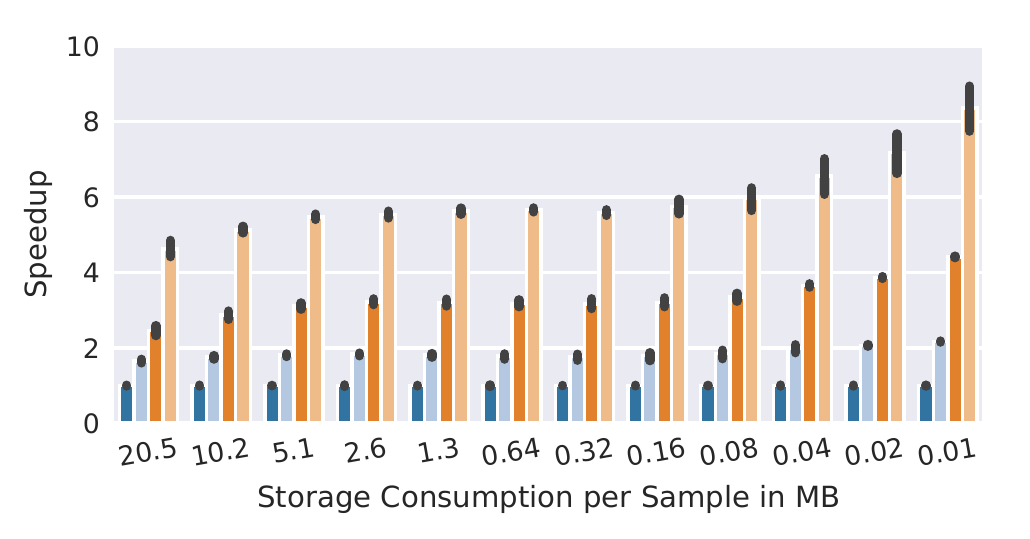}
        \vspace{-18pt}
        \caption{TensorFlow RMS}
        \label{fig:synthetic-dataset-processing-tf}
    \end{subfigure}
    \vspace*{-0.3cm}
    \caption{{\color{diff2}Speedup of applying RMS to a synthetic 15\:GB dataset with different sample sizes implemented in NumPy and TensorFlow. }}
    \label{fig:synthetic-dataset-tf-vs-np}
\end{figure}
\vspace{-0.3cm}

To test how external libraries affect the throughput, we created a new preprocessing step which applies the root-mean-square (RMS) function with a period of 500 over the entire sample, leaning on a similar computation from the NILM pipeline.
We implemented this step both with NumPy and TensorFlow, and we profile our synthetic datasets with steps applied online individually.
The results in Fig.\ref{fig:synthetic-dataset-tf-vs-np} show that the NumPy implementation has the same slowdown as NILM and NLP for all sample sizes, whereas the TensorFlow implementation shows a speedup between 4-8$\times$ for eight threads.
However, while the NumPy implementation does not scale, it is still 2.9$\times$ faster with a single-threaded processing time of 650 seconds compared to TensorFlow's 1905 seconds with eight threads at the 20.5\:MB sample size.
In other words, it pays off to use the less scalable but more efficient implementation in NumPy instead of the native implementation in TensorFlow.

}

{\color{diff}

\textbf{(3) Random file access performance can affect the speedup. }
We have already discussed the impact of concatenation in Sec.~\ref{ssec:storage-versus-throughput} \textbf{(1)} and how random file access can hinder achieving high throughput and bandwidth utilization.
By running the multi-threading experiments with system-level caching enabled, we can isolate the effect of random file access on speedup.
For example, the \texttt{unprocessed} strategy of the MP3 pipeline (Fig.~\ref{fig:speedup-mp3}) has a speedup of $2\times$ when reading from storage with eight threads, versus a speedup of $6\times$ when reading from memory (Fig.~\ref{fig:speedup-epochs-mp3}).
This shows that decoding does in fact scale well.
The same effect changes the speedup of the FLAC pipeline from 4$\times$ (Fig.~\ref{fig:speedup-flac}) to 6$\times$ with eight threads (Fig.~\ref{fig:speedup-epochs-flac}).

}

\subsection{Shuffling}
\label{ssec:shuffling}

A common technique in DL is to change the order of the dataset in every epoch, so the optimizers do not see the same gradients in mini-batches.
There are a few different approaches to shuffling the dataset, which include sampling from the dataset \textit{with-} or \textit{without replacement}~\cite{de2020random,haochen2019random,yun2021singleshuffle}.
Irrespective of which algorithm is used to modify the order of the dataset, it is a very memory-intensive problem, as the entire dataset has to be loaded into RAM.
One solution is to create a buffer that fits into the memory and use a \textit{with-replacement} sampling strategy to iterate over the entire dataset in a pseudo-random fashion~\cite{tfdatasetapishuffle}, similar to reservoir sampling~\cite{10.1145/3147.3165}.

We implemented and profiled this approach with multiple sample counts, which confirms the naive assumption that the \textit{per-sample} processing time for shuffling is constant.
That means that shuffling has a linear relation to storage consumption and is not specific to a pipeline or dataset.
The difference in per-sample processing time between shuffling and not shuffling for each sample size is $(\pm0.5)$ 9.6ms on average.
An additional characteristic is that the initial call to allocate a buffer is amortized with a bigger sample size, which manifests itself in the increasingly faster per-sample time with incremented sample sizes. 

PRESTO's profiling can help to find the optimal place in the pipeline where shuffling should be applied.
As the storage consumption of a strategy does not affect the runtime of shuffling, we do not recommend making shuffling part of the strategy selection.
However, once a strategy is determined, we suggest to shuffle after the online pipeline step that yields the smallest data size.
If we consider a fixed-size buffer for shuffling, the highest number of samples can be fit into the buffer when the size of the data sample is smallest.
The higher the number of samples in the buffer, the higher the entropy; this, in turn, leads to a better approximation of the ``true'' gradient~\cite{ruder2016overview,kingma2014adam}.

{\color{diff}

\subsection{Modifying the Pipeline}
\label{ssec:modifying-pipeline}

We added an additional preprocessing step to the CV pipeline to showcase how the trade-offs can shift in an already profiled pipeline.
We decided on adding a step that converts images from RBG to greyscale because this is a common preprocessing step that affects the storage consumption and is not obviously compute intensive.
To evaluate how an additional step will affect the pipeline performance, we profiled two setups: before and after the \texttt{pixel-centered} strategy.

}

\vspace{-0.2cm}
\begin{figure}[h]
    \begin{subfigure}[c]{0.23\textwidth}
        \includegraphics[width=\textwidth]{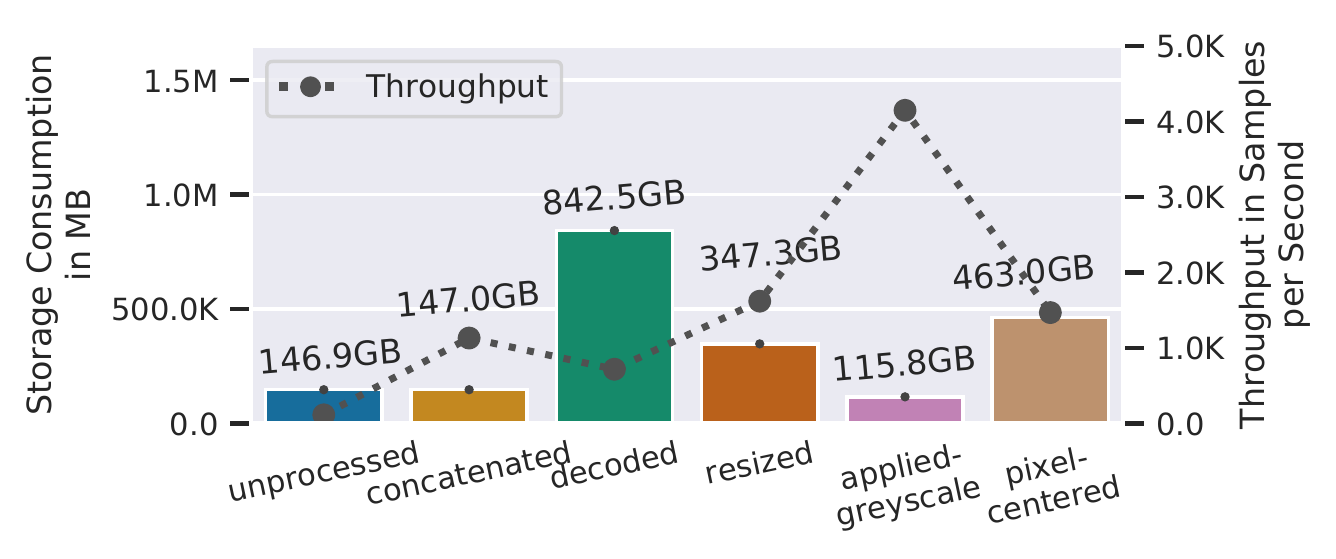}
        \vspace{-18pt}
        \caption{{\color{diff}Before Pixel Centering}}
        \label{fig:ss-vs-thr-gs-before-pc}
    \end{subfigure}
    \begin{subfigure}[c]{0.23\textwidth}
        \includegraphics[width=\textwidth]{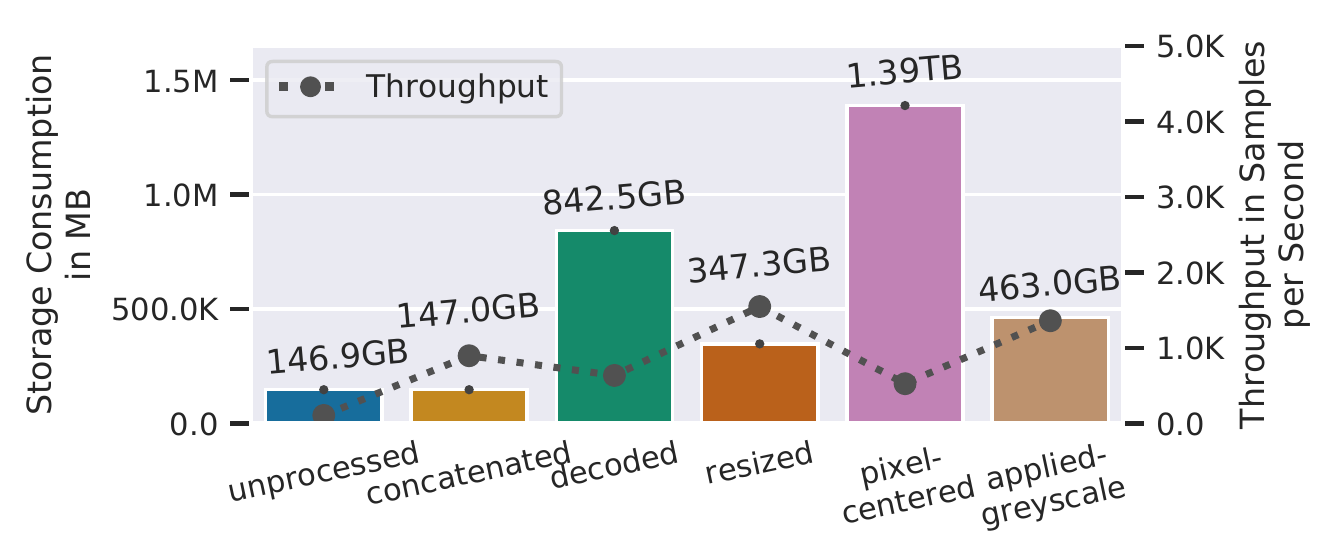}
        \vspace{-18pt}
        \caption{{\color{diff}After Pixel Centering}}
        \label{fig:ss-vs-thr-gs-after-pc}
    \end{subfigure}
    \vspace{-10pt}
    \caption{{\color{diff}Storage consumption (left y-axis) and throughput (right y-axis, dotted line) comparison of adding a greyscale transformation before and after the pixel centering.}}
    \label{fig:imagenet-greyscale-comparison}
\end{figure}
\vspace{-0.2cm}

{\color{diff3}

Before discussing the results, we explain the characteristics of the new \texttt{applied-greyscale} strategy and compare it to \texttt{resized}.
Converting a 3-channel image to greyscale should decrease the storage consumption by 3$\times$, because we only need a single channel with the same datatype.
The \texttt{resized} strategy reduces the size by 2.4$\times$ for the CV dataset, which is dependent on the average image resolution.
Adding greyscaling will also affect the throughput of \texttt{pixel-centered} as the final storage consumption will be reduced as well.

The results in Fig.~\ref{fig:imagenet-greyscale-comparison} show the effect of the additional greyscale step on the storage consumption and throughput. 
First of all, applying the greyscale step before the \texttt{pixel-centered} strategy increases the maximum throughput of the pipeline by 2.8$\times$, from 1513\:SPS with \texttt{resized} (Fig.~\ref{fig:ss-vs-thr-gs-after-pc}) to 4284\:SPS with \texttt{applied-greyscale} (Fig.~\ref{fig:ss-vs-thr-gs-before-pc}).
While the last strategy of both setups performs similary, applying preprocessing steps that reduce the storage consumption consecutively increases the throughput for intermediate strategies.
We additionally evaluated a setup where the resize and greyscale steps are interchanged before the pixel centering, but there was no significant difference in performance to Fig.~\ref{fig:ss-vs-thr-gs-before-pc}.
The second setup with the last strategy \texttt{applied-greyscale} increases the throughput from 534\:SPS at \texttt{pixel-centered} to 1384\:SPS by reducing the data size from 1.4\:TB to 463\:GB.
This supports our observation \textbf{(2)} from Sec.~\ref{ssec:storage-versus-throughput} that steps which reduce storage consumption should be investigated with priority when searching for the best performing.

}



\vspace{-0.5cm}
\section{Lessons Learned}
\label{sec:lessons-learned}




We find it important to summarize our findings more generically for both DevOps and ML practitioners so that they can use the PRESTO library and the generated insights for future analysis. These lessons are based on the analysis in Section~\ref{sec:analysis}.

\textbf{(1) Storage consumption is an important characteristic when estimating throughput. }
We gravely underestimated the effect of storage consumption before conducting this study.
The impact of storage consumption is a multi-faceted one, as it affects the storage hardware, its interconnects, the deserialization process and multi-threading capabilities.
{\color{diff2} 
A small total storage consumption performs best if not throttled by a CPU bottleneck, and steps that reduce data size should be prioritized when searching for the best performing strategy.
However, small sample sizes ($\leq 0.08$\:MB) increase the online processing time dramatically irregardless of reading from storage or from memory and can be a reason for an underutilized I/O bandwidth.
These two observations combined are the reason why fully preprocessed datasets did not yield the best throughput in 4 (CV, CV2-PNG, CV2-JPG, NLP) out of 7 pipelines.
}

{\color{diff2} 
\textbf{(2) Multi-threading usually improves throughput, but the speedup can be limited for various reasons. }}
Parallel execution of a pipeline is not a silver bullet when trying to speed up preprocessing.
{\color{diff2}
First of all, various issues can impede parallel speedup, such as calling external Python libraries or dealing with extremely short-running preprocessing tasks at small sample sizes.
But even when parallel speedup of a strategy is reasonably good, a different strategy with a lower data volume to be read from storage may perform much better.
}

{\color{diff2} 
\textbf{(3) It is recommended to use application-level caching whenever possible. }
Whenever the dataset fits into memory, application-level caching increased the throughput in our experiments by up to 15$\times$ with a high sample size.
Application-level caching improved the throughput compared to system-level caching by a factor of 1.3-4.6$\times$, and should be preferred as the deserialization of cached files can slow down the pipeline.
}

{\color{diff2} 
\textbf{(4) Compression can be useful when not facing a CPU bottleneck. }
Compression can increase the throughput by a factor of 1.6-2.4$\times$ under few conditions: a high enough space saving of 73-93\% and the absence of computationally expensive processing steps.
However, estimating the space saving, as well as the decompression time is hard.
Additionally, applying compression can increase the offline processing time between 1.1$\times$ and 13.5$\times$ compared to no compression.
The overheads of compression should be taken into account and carefully weighted against I/O savings.
}

\vspace{-0.8cm}
\section{Related Work}
\label{sec:related-work}
I/O profiling was already done in a micro-benchmark for TensorFlow by Chien et al.~\cite{Chien_2018} which focused on different file systems and the check-marking functionality.
In their experimental setup with AlexNet~\cite{krizhevsky2012imagenet}, the training data prefetching eliminated the effective preprocessing time, similar to our CV pipeline with the \texttt{unprocessed} strategy.

OneAccess, a unified data loading layer, functions as middleware for preprocessing and helps to run ML jobs on multiple nodes more efficiently by removing duplicate processing for hyperparameter tuning~\cite{kakaraparthy2019case}.
We have observed similar results where packing the dataset helps by allowing sequential data access, but their preprocessing seems to be done fully offline.
Their sample lifecycle concept, which plans to store the data for a certain amount of time in anticipation of re-use, is a perfect fit for PRESTO's strategy optimization to select the lowest storage consuming dataset representation.

An example of improved I/O efficiency in DL for high-performance computing (HPC) is the framework DeepIO by Zhu et al.~\cite{zhu2018entropy} which optimizes data loading to improve the training throughput.
This framework could be used to complement our methodology to mitigate I/O bottlenecks.

Model throughput, which is coupled with GPU utilization, has been identified as an essential topic in the MLOps community~\cite{Jeon2018,ozeri2018object,murray2021tf}.
Microsoft pointed out in a study~\cite{Jeon2018} that underutilization of GPUs in multi-tenant settings is a problem for cloud providers.
They evaluated how job locality and human errors can lead to unnecessarily idling resources.
IBM implemented their FUSE-based~\cite{fuse2018} file system for object storage to improve the I/O loads in their IBM Fabric for Deep Learning services~\cite{ozeri2018object}.
Additionally, they deployed caching mechanisms to improve long-term read throughputs if the dataset fits into memory.
These studies touch on different pain points of the cloud providers as they start to recognize the potential of improving the deployment and resource usage of end-to-end DL pipelines, which integrate the previously overlooked preprocessing phase.

Another work in that direction is a recent NVIDIA study~\cite{nvidiabenchmarks2020} which profiled the end-to-end training and inference time for multiple frameworks and models on their GPU and TPU servers. They highlight the benefits of different hardware solutions but do not take the preprocessing pipeline into account as they preprocess the dataset only once completely.

Relocating the preprocessing to more specialized hardware can increase performance and adds additional resources to the preprocessing profiling.
NVIDIA's Data Loading Library (DALI), a Python library~\cite{dali2020}, provides common preprocessing steps for images, video, and audio formats, which can be used as a drop-in replacement for native pipelines that can be executed on the GPU.
DALI has shown to improve the performance of multiple end-to-end DL pipelines~\cite{mohan2020analyzing,dalibench2019}.
PRESTO can be applied on a DALI-enhanced pipeline, and while this may shift the trade-offs, it is essential to note that additional resources also introduce complexities like bandwidth restrictions, limited memory sizes, and in this case, double-use for preprocessing as well as training.
DL models are stored in GPU memory for forward and backward passes, interfering with the improved preprocessing execution due to restricted memory size and computational capabilities when executed in a pipelined fashion.
The cost and ubiquity of CPU processing should be weighed carefully against GPUs and when in doubt, be optimized in an end-to-end manner with tools like CoorDL~\cite{mohan2020analyzing}.

Our work focused on the trade-offs between storage consumption, throughput, and preprocessing time, while Mohan et al.~\cite{mohan2020analyzing} analyzed different types of stalls and focused on dividing the entire end-to-end DL pipeline into data fetches, preprocessing rate, and GPU processing rate.
They have shown how to efficiently use the OS-level cache to improve fetch stalls while increasing the total time-to-accuracy on two 24 core machines with 8 GPUs and 500\:GB RAM with an HDD and an SSD local storage.
While our focus was more on consumer-level hardware, which does not allow this level of caching, we had a similar observation that the decoding step in CV is very inefficient and that slow preprocessing can bottleneck the training performance on the GPU.
We provide a solution to one of their discussion points on mitigating the increased storage consumption due to decoding with a suitable strategy, which can cache the entire dataset even more efficiently combined with CoorDL.

Relocating the preprocessing onto an accelerator is also done in SMOL, a system that prepares the most efficient strategy on how to preprocess visual data \textit{and} train a model in an end-to-end fashion while keeping the accuracy fixed~\cite{kang2020jointly}.
We reproduced similar preprocessing bottlenecks regarding our image pipeline.
However, while SMOL uses data compression steps and other techniques to speed up the data processing while providing the same model accuracy, we focused entirely on the preprocessing pipeline.
Some of our insights can be incorporated into SMOL, {\color{diff2} such as partially preprocessing a pipeline for a specific set of hardware, allowing better throughput based on the presence of hardware en-/decoders or having additional compression in the preprocessing pipeline to increase the final throughput.}
Supplementing SMOL with our analysis of common preprocessing steps could enable it to work on non-image data.

A very recent publication, Plumber~\cite{kuchnik2021plumber}, focuses on online, automatic optimization of \texttt{tf.data} pipelines through the use of \textit{resource accounted rates} that estimate the CPU, disk and memory requirements. We share insights like the benefits of caching and unexpected parallelization performance from user-defined-functions. However, our work proposes a new perspective at the available trade-offs in preprocessing pipelines by statically analyzing them, while Plumber focuses on the parallelization level, prefetch buffer sizes, and configuration of the \texttt{tf.data} building blocks to overcome I/O inefficiencies.
\vspace{-0.2cm}
\section{Discussion}
\label{sec:discussion}
While our analysis provides some key insights about how to profile and configure a preprocessing pipeline, we want to highlight some settings which could benefit from further research.

\textbf{Datasets can grow over time.}
The results from PRESTO when profiling a pipeline and a static dataset should provide valuable insights if the dataset grows in the future.
One exception is when the data representation of the newly added data is not compatible with the previous dataset, e.g., adding 4k images to a VGA-resolution dataset, which may slow down parts of the pipeline in unpredicted ways and result in different trade-offs.
TensorFlow Extended (TFX)~\cite{10.1145/3097983.3098021} is an ML platform to train and deploy models, which can be used to keep track of this shift in the dataset.

\textbf{Storage bandwidth has shown to be a bottleneck for other similar MLOps studies~\cite{mohan2020analyzing, murray2021tf, kang2020jointly}}.
{\color{diff2}We have shown that compression is a promising tool to mitigate storage-related bottlenecks but its efficacy is limited. Compression that is optimized to store \textit{tensor-like} data could potentially provide even better throughput and space saving.}
Our recommended strategies from CV and NLP are integer tensors, but NILM, MP3, and FLAC use floating-point tensors with 32 and 64 bit, which suggests that different compression algorithms have to be considered depending on the data representation~\cite{lemire2015decoding, fastfpd}.
When applied carelessly, compression can have severe effects on the entire online processing.
PRESTO can be used to study the effect of compression in more depth.


\textbf{Distributed computing for preprocessing.}
A common solution to speed up the execution jobs is using multiple worker nodes with frameworks like Apache BEAM~\cite{beam} or Spark~\cite{zaharia2010spark}.
Preprocessing a dataset is a trivially parallelizable task by splitting the dataset into equal chunks for every worker to process simultaneously, except for shuffling or similar global dataset operators.
While it is easy to follow PRESTO's recommendation and apply the offline transformation steps until the desired data representation is met, there are more complexities involved, like the data locality to workers, the locality of the workers to the training process, the amount of workers available, the interconnects, and the scheduling algorithm that supervises the job execution.
This distributed setting will benefit from PRESTO's analysis, as storage consumption is correlated with the network bandwidth usage, and finding a strategy that has a good speedup will be even more effective with multiple workers. 
These insights may help to improve data management and scheduling.
Nevertheless, additional profiling should be done to further optimize the pipeline execution for the specific cluster-computing framework.

{\color{diff}
\textbf{Applicability for concurrent training.}
When considering a setup with a shared preprocessing pipeline between multiple distributed training jobs, such as in hyperparameter tuning, all of our insights are applicable, as the throughput $T_4$ can be fanned out to all training jobs. However, this setup adds load onto the network between the preprocessing node and the training nodes, which would not happen when running the preprocessing locally on the same machine that performs training. If the network can not handle the duplicated load of fanning out the preprocessed data per training job, it will become a new bottleneck.
}

\vspace{-0.3cm}
\section{Conclusions}
\label{sec:conclusion}
This paper presents an analysis of seven concrete DL pipelines based on their typical preprocessing steps from CV, NLP, NILM, and the Audio domain.
We provide a profiling library, PRESTO, that helps with detecting bottlenecks and automatically decide which preprocessing strategy is the most efficient based on an {\color{diff2}objective function}.
We show that not preprocessing the dataset before training is never the best solution for all pipelines, and fully preprocessing can affect the final preprocessing throughput negatively due to problems relating to I/O and storage consumption.
Alternatively, we propose different strategies that increase the CV pipeline throughput by {\color{diff2}$3\times$ and NLP by $13\times$} while reducing their storage consumption compared to the fully preprocessed dataset.
We provide insights into how storage consumption, {\color{diff2}different caching level and compression affect} the preprocessing pipeline and how they can pinpoint where bottlenecks are formed.
While multi-threading could be an effective way to speed up preprocessing, we show that using an intermediate preprocessing strategy is significantly more impactful to reduce processing time.
Finally, we provide an intuition about profiling the preprocessing pipelines effectively by summarizing the generated insights to mitigate future bottlenecks in deep learning pipelines.


\paragraph*{Acknowledgements}
This work is funded in part by the Deutsche Forschungsgemeinschaft (DFG, German Research Foundation) - 392214008.

\bibliographystyle{ACM-Reference-Format}
\bibliography{main}










\end{document}